\documentclass[fleqn,12pt]{wlscirep}

\usepackage[utf8]{inputenc}
\usepackage[T1]{fontenc}
\usepackage{titling}
\usepackage{bm}
\usepackage{setspace}
\usepackage{caption}
\usepackage{graphicx}
\usepackage{subcaption}
\usepackage{multirow}
\usepackage{parskip}
\usepackage{amsthm}
\usepackage{mathrsfs}
\usepackage{textcomp}
\usepackage{manyfoot}
\usepackage{tabularx}
\usepackage{array}
\usepackage{algorithm}
\usepackage{algorithmicx}
\usepackage{algpseudocode}
\usepackage{listings}
\usepackage{url}
\usepackage{placeins}
\usepackage{orcidlink}

\newcolumntype{Y}{>{\raggedright\arraybackslash}X}
\newcommand{\botrule}{\bottomrule}

\begin{document}
\hypersetup{pageanchor=false}

\begin{titlepage}
\raggedright
\textbf{Review Article} \\
\vspace{1.0cm}

\begin{center}
\LARGE\textbf{
Medical world models in healthcare: capabilities, evidence, and trustworthy clinical translation} \\
\vspace{1.2cm}
\small
\textbf{Wanshu Fan\textsuperscript{1}~\orcidlink{0000-0001-6299-2795},
Zhongxiu Cong\textsuperscript{1},
Zhaoyan Chen\textsuperscript{1}
Zhuanfeng Jin\textsuperscript{1},
Qi Ai\textsuperscript{2},
Haifan Gong\textsuperscript{3},
Congyu Liao\textsuperscript{4},
Xiaofeng Liu\textsuperscript{5},
Dongsheng Zhou\textsuperscript{1,*}~\orcidlink{0000-0003-3414-9623},
Cong Wang\textsuperscript{6,*}~\orcidlink{0000-0002-6068-0103}} \\
\vspace{0.3cm}

\textsuperscript{1}National and Local Joint Engineering Laboratory of Computer Aided Design, School of Software Engineering, Dalian University, Dalian, China\\
\textsuperscript{2}Department of Radiology, Xinhua Hospital Affiliated to Dalian University, Dalian, China\\
\textsuperscript{3}The Chinese University of Hong Kong, Shenzhen, China\\
\textsuperscript{4}University of California, San Francisco, USA\\
\textsuperscript{5}Yale University, New Haven, USA\\
\textsuperscript{6}The Hong Kong Polytechnic University, Hong Kong, China
\end{center}

\vspace{0.3cm}
{Correspondence and requests for materials should be addressed to:\\

Dongsheng Zhou: zhouds@dlu.edu.cn\\
Cong Wang: supercong94@gmail.com}

\vfill
\end{titlepage}
\hypersetup{pageanchor=true}

\newpage
\raggedright
\textbf{\large Abstract}
\vspace{1cm}

Medical world models offer a framework for extending medical artificial intelligence beyond static prediction by representing evolving patient states and modelling how they change over time and in response to clinical interventions. This Review provides a structured narrative synthesis with reproducible evidence mapping to define the conceptual boundaries, technical foundations, application domains, and clinical evidence requirements of the field. 
We screened 1,455 unique records and assembled a corpus of 300 sources, including 14 studies that met a strict empirical definition of a medical world model. The field is organized around four capabilities: patient state representation, temporal dynamics modelling, intervention-conditioned simulation, and clinician-supervised planning. Evidence spans medical imaging, longitudinal electronic health records, treatment response modelling, physiological and multimodal state modelling, ultrasound and surgical interaction, and population and health-system simulation; clinical digital twins are treated as a cross-cutting integration framework. 
Current studies provide early evidence of technical feasibility for trajectory forecasting and comparison of candidate interventions, but most remain retrospective, task-specific, or preclinical. Importantly, we distinguish functional capability from clinical evidence maturity and specify the evidence required to support progressively stronger clinical claims, from temporal forecasting to intervention comparison and planning-oriented decision support. 
These requirements include precise intervention representations, robust causal and mechanistic grounding, calibrated trajectory-level uncertainty, safety-constrained planning, external validation, and prospective multicentre evaluation against clinically meaningful endpoints. 
This evidence-oriented framework is intended to help distinguish technically plausible simulation from clinically supportable prediction and decision support.

\vspace{2cm}

\newpage
\setstretch{1.2}
\section{Introduction}\label{sec:introduction}

Medical artificial intelligence (AI) has progressed from rule-based systems and task-specific prediction models towards large-scale models capable of learning transferable representations across tasks, modalities, and clinical settings. Deep learning, foundation models, and generative AI have substantially advanced medical image analysis, electronic health record (EHR) modelling, multimodal data integration, and clinical decision support \cite{ref026,ref031,ref085,ref101,ref102}. Nevertheless, most current systems remain designed to infer a diagnosis, risk estimate, or outcome from observations available at a particular time point. 
Clinical decision-making is inherently longitudinal and intervention-dependent: clinicians must determine not only a patient's current condition, but also how that condition is likely to evolve, how alternative interventions may alter its course, and which management strategy offers the most favourable balance of benefit and risk. 
Such questions cannot be adequately addressed using isolated images, individual EHR entries, or single laboratory measurements, particularly when disease trajectories are prolonged, heterogeneous, and only partially observed \cite{ref013,ref065,ref098,ref099}.

World models offer a computational framework for representing and simulating dynamic processes. Developed primarily in model-based reinforcement learning, they learn compact representations of an environment together with transition dynamics that describe how its state evolves over time and, in more advanced settings, in response to actions. These learned dynamics can be used to generate plausible future states, evaluate candidate actions, and support planning through internal simulation, thereby reducing the need for repeated interaction with the real environment \cite{ref001,ref002,ref005,ref017,ref028,ref082}. Recent advances in latent state-space modelling, self-supervised predictive learning, diffusion and video generation, and embodied AI have extended world-model research beyond conventional control tasks towards increasingly complex, multimodal, and partially observed environments \cite{ref028,ref094,ref034,ref053,ref087,ref128}.

In medicine, world-model principles recast patients and care processes as evolving systems rather than collections of independent observations. A medical world model may integrate medical images, longitudinal EHR data, physiological signals, laboratory measurements, multi-omics data, and clinical events into a representation of the patient's current state, and then model how that state changes over time. At a basic level, such models predict subsequent latent states or future clinical observations. More advanced systems explicitly condition state transitions on treatments, procedures, acquisition parameters, or device movements, allowing them to simulate intervention-dependent trajectories and, in some cases, compare candidate actions. When combined with appropriate causal assumptions, uncertainty estimation, safety constraints, and clinician oversight, these capabilities may support disease-course forecasting, treatment-response simulation, counterfactual comparison, and planning-oriented clinical decision support \cite{ref056,ref095,ref047,ref039}. Importantly, temporal prediction, action-conditioned simulation, counterfactual comparison, and closed-loop planning represent distinct levels of functionality and should not be treated as interchangeable capabilities.

Early studies highlight both the breadth of medical world models and the nascent state of their evidence base. EHRWorld models long-horizon trajectories from longitudinal clinical records; EchoJEPA and EchoWorld investigate predictive representations and probe-conditioned transitions in echocardiography; CLARITY and MeWM simulate treatment-conditioned disease evolution and compare candidate treatment strategies; and Brain-WM jointly predicts treatment sequences and future glioblastoma imaging \cite{ref047,ref070,ref074,ref046,ref039,ref050}. Related work extends these concepts to medical-image representation, physiological-state modelling, surgical interaction, and simulation at the population and health-system levels. However, most studies remain retrospective, task-specific, or preclinical, and technical capability should not be conflated with clinical effectiveness. Conceptual boundaries also remain unsettled: the term ``medical world model'' has been applied to predictive representation models, longitudinal patient simulators, treatment-conditioned generative systems, embodied control models, and clinical digital twins, despite substantial differences in state definition, action semantics, prediction horizon, and intended use. Moreover, general world models, medical foundation models, causal treatment-effect modelling, and digital twins have largely been reviewed in isolation, leaving the scope, capability hierarchy, and clinical evidence requirements of medical world models insufficiently defined.

Several developments make this an opportune moment to examine medical world models critically. First, longitudinal health data are now available at a scale that supports generative modelling of entire disease histories. Delphi-2M, a generative transformer trained on 0.4 million UK Biobank participants, predicts the rates of more than 1,000 diseases conditional on past history, was validated on 1.9 million Danish individuals without parameter changes, and can sample synthetic health trajectories for up to 20 years \cite{shmatko2025delphi}; related timeline and clinical-language models show that tokenized health records and clinical notes can serve as general-purpose substrates for trajectory prediction \cite{renc2024ethos,jiang2023nyutron,ref041}. Second, foundation models for pathology, ophthalmology, radiology, and clinical language now provide pretrained encoders from which patient states can be constructed \cite{moor2023foundation,chen2024uni,xu2024gigapath,ref103,singhal2025medpalm2}. Third, general-purpose world models have matured from game environments to video-scale physical simulators and scalable latent planners \cite{ref016,hansen2024tdmpc2,assran2025vjepa2,nvidia2025cosmos}, and these architectures are being adapted to surgical, ultrasound, and endovascular robotics through large multi-institutional datasets \cite{openh2026embodiment,he2025cosmoshsurgical}. Fourth, digital twins have attracted national-level attention in precision medicine, and a recent consensus report from the US National Academies identified verification, validation, and uncertainty quantification as foundational research needs for their use in decision making \cite{nasem2024digitaltwins,venkatesh2022healthdt,ref110}. These trends create both opportunity and risk: the generative capacity that enables realistic simulation also makes it easier to produce trajectories that appear plausible but are not clinically valid.

Existing reviews approach this territory from different directions (Table~\ref{tab:prior_reviews}). General world-model surveys synthesize architectures, reasoning paradigms, and embodied applications, but address medicine only briefly \cite{ref028,ref034,ref087,ref094,ref128}. Reviews of medical foundation models focus on transferable representations rather than on intervention-conditioned dynamics \cite{ref026,ref031,moor2023foundation}. Digital-twin reviews emphasize mechanistic simulation, data integration, and synchronization with the physical counterpart \cite{ref110,ref117,ref044,zhou2026digitaltwinai}, whereas causal machine-learning reviews formalize treatment-effect estimation without addressing generative rollout or embodied control \cite{ref013,ref065,feuerriegel2024causal}. A small but growing set of perspectives has begun to articulate medical, biomedical, population-health, and epidemiological world models directly \cite{ref043,ref095,ref056,wang2026biomedwm,xie2026phwm,memon2026epiwm,feng2026dynamictraj}, and domain-specific surveys now cover surgical video generation and robotic ultrasound \cite{huang2026surgvideosurvey,jiang2023roboticus}. This Review differs from these works in combining a reproducible evidence map with an explicit separation between functional capability and clinical evidence maturity, and by specifying the evidence required for each class of clinical claim.

\begin{table}[!t]
\footnotesize
\setlength{\tabcolsep}{3pt}
\renewcommand{\arraystretch}{1.12}
\caption{Positioning of this Review relative to related reviews and perspectives. Categories are grouped by primary focus; individual reviews may span more than one category. The third column summarizes the emphasis of each category, and the fourth column describes the distinctive emphasis of the present Review rather than a deficiency of prior work.}\label{tab:prior_reviews}
\begin{tabularx}{\textwidth}{@{}>{\raggedright\arraybackslash}p{0.19\textwidth}>{\raggedright\arraybackslash}p{0.19\textwidth}YY@{}}
\toprule
Category & Representative sources & Primary focus & Distinctive emphasis of this Review\\
\midrule
General world-model surveys & \cite{ref028,ref034,ref087,ref094,ref128,ref053} & Architectures, video generation, embodied and autonomous-driving applications, reasoning paradigms & Medical state and action semantics; separation of functional capability from clinical evidence\\
Medical foundation-model reviews & \cite{ref026,ref031,moor2023foundation,acosta2022multimodal} & Large-scale pretraining, multimodal representation, generalist tasks & Temporal and intervention-conditioned dynamics built on pretrained encoders\\
Digital-twin reviews and reports & \cite{ref110,ref117,ref044,venkatesh2022healthdt,nasem2024digitaltwins,zhou2026digitaltwinai} & Mechanistic and hybrid simulation, synchronization with the physical counterpart, verification and validation & Learned latent dynamics, generative rollout, and capability-specific evidence requirements\\
Causal machine-learning and trajectory reviews & \cite{ref013,ref065,feuerriegel2024causal,feng2026dynamictraj} & Identification and estimation of individualized and longitudinal treatment effects & How causal estimands constrain generative rollout, planning, and embodied control\\
Medical and biomedical world-model perspectives & \cite{ref043,ref095,ref056,wang2026biomedwm,xie2026phwm,memon2026epiwm} & Conceptual agendas, capability taxonomies, and opportunities across clinical, biomedical, population, and epidemiological settings & Reproducible evidence mapping, a strict empirical subset, coverage of recent studies, and matching of evidence to claims\\
Domain-specific surveys & \cite{huang2026surgvideosurvey,ref044,jiang2023roboticus} & Surgical video generation, surgical digital twins, robotic ultrasound & Cross-domain comparison of state, action, and validation across six application domains\\
\botrule
\end{tabularx}
\end{table}

In this Review, we use a structured narrative approach with reproducible evidence mapping to define the conceptual and functional boundaries of medical world models and position them within the broader evolution of medical AI. We organize the field around four core capabilities: patient-state representation, temporal dynamics modelling, intervention-conditioned simulation, and clinician-supervised planning. 
The literature is synthesized across six overlapping application domains: medical imaging and future-observation simulation; longitudinal EHR and patient-trajectory modelling; disease-progression and treatment-response simulation; physiological-signal and multimodal-state modelling; ultrasound guidance and surgical embodied interaction; and population-level and health-system simulation. Clinical digital twins are treated as a cross-cutting integration framework rather than a separate application domain. Beyond mapping the field, we distinguish functional capability from clinical evidence maturity and establish an evidence-oriented framework that specifies the requirements for supporting progressively stronger clinical claims, from temporal forecasting and intervention-conditioned simulation to alternative-intervention comparison and planning-oriented decision support. These requirements encompass longitudinal data integrity, explicit intervention semantics, causal validity, calibrated uncertainty, external and prospective validation, clinically meaningful endpoints, safety, and governance.

The Review makes five contributions. First, it delineates medical world models from adjacent paradigms, including clinical prediction models, medical foundation models, generative models, causal treatment-effect models, mechanistic simulators, clinical digital twins, and reinforcement-learning policies, using explicit criteria based on state, action, transition, and rollout (Table~\ref{tab:concepts}). Second, it provides a unified probabilistic formulation that links patient-state inference, intervention-conditioned transitions, constrained planning, and the causal estimands that rollouts may or may not identify. Third, it synthesizes the literature across six application domains and maps representative studies, including recent studies, to capability levels and evidence boundaries. Fourth, it relates each class of clinical claim to the data, causal, uncertainty, validation, and governance evidence required to support it, together with common threats to validity and corresponding diagnostic checks. Fifth, it proposes a research agenda for causal--mechanistic, uncertainty-aware, and clinically governed medical world models. Section~\ref{sec:methods} describes the review methods; Section~\ref{sec:history} traces the development of medical AI; Section~\ref{sec:framework} presents the conceptual and technical framework; Section~\ref{sec:applications} reviews applications, evidence, and translational requirements; Section~\ref{sec:future} outlines open problems and future directions; and Section~\ref{sec:conclusion} concludes.

\section{Methods}\label{sec:methods}

\subsection{Review Design and Scope}

We conducted a structured narrative review with reproducible evidence mapping to define the conceptual boundaries of medical world models, characterize their technical architectures and application domains, compare their functional capabilities using an L1--L4 framework, and distinguish technical performance from clinical evidence maturity. The evidence-mapping strategy drew on established scoping-review methodology and reporting guidance \cite{arksey2005scoping,tricco2018prismascr}, while the reporting of study identification and electronic searches was guided by relevant elements of PRISMA 2020 and PRISMA-S \cite{page2021prisma,rethlefsen2021prismas}. Because the final corpus included purposively selected foundational, contextual, methodological, and reporting sources in addition to empirically identified studies, this Review is not presented as a systematic review and does not claim full compliance with PRISMA-ScR. No protocol was prospectively registered; however, a dated internal protocol, complete search logs, and a record of protocol deviations were retained.

\subsection{Information Sources and Search Strategy}

Electronic searches were conducted in PubMed, arXiv, IEEE Xplore, and the
ACM Digital Library for records dated from 1 January 2018 to the
platform-specific final search date. PubMed, arXiv, and IEEE Xplore were
searched on 19 July 2026, and the ACM Digital Library was searched on 20 July
2026.

Two complementary query families were used. The explicit world-model query
combined (``world model'' OR ``world models'' OR ``medical world model'' OR
``clinical world model'' OR ``patient world model'') with a medical-context
block comprising medical, clinical, patient, health, healthcare, electronic
health record, EHR, disease, imaging, radiology, ultrasound, surgery,
physiological, and hospital. The capability query replaced the first block
with (``latent dynamics'' OR ``state transition model'' OR
``action-conditioned simulation'' OR ``patient trajectory simulation'' OR
``disease trajectory simulation'' OR ``model-based reinforcement learning''
OR ``internal rollout''). Syntax, field restrictions, and other
platform-specific adaptations are reported in the Supplementary Information.

The searches retrieved 612 records from PubMed, 663 from arXiv, 273 from
IEEE Xplore, and 1,959 from the ACM Digital Library. Twenty-one candidate
records identified during preliminary manuscript development were included
as an initial seed set. These database records and seed records constituted
the initial identification set subsequently processed as described in
Section~\ref{sec:record-processing}.

Directed sources were identified throughout manuscript development. These included targeted arXiv searches combining ``world model'' with medical, clinical, patient, surgical, ultrasound, endoscopic, radiological, and electronic-health-record terms, and purposive identification of foundational, methodological, data-resource, evaluation, and governance literature. Directed sources were handled as described in Section~\ref{sec:record-processing}.

\subsection{Eligibility Criteria and Corpus Roles}

For inclusion in the strict empirical subset, studies were required to
address a medical or health-related setting, empirically evaluate a learned
model of dynamic state evolution rather than merely predict an isolated
future outcome, and demonstrate at least one operational world-model
capability: representation of an evolving patient or environment state,
learned transition dynamics, action- or treatment-conditioned simulation,
multi-step internal rollout, or interaction within a learned dynamic
environment.

Studies limited to static classification, segmentation, detection, question
answering, or generation without explicit temporal or action-related
semantics were excluded from the strict subset. Studies that predicted only a
single future label or endpoint without modelling state evolution or learned
transitions were likewise excluded. Longitudinal representation-learning
studies were not automatically classified as medical world models unless they
implemented and evaluated a qualifying dynamic mechanism. Conventional
health-economic Markov models and other predefined state-transition analyses
were excluded unless they implemented and empirically evaluated a learned
patient-level state-transition or rollout mechanism comparable to an
operational world model.

Both peer-reviewed publications and preprints were eligible. When a preprint
and a subsequently published article reported the same study, they were
merged and treated as one study, with the most complete version used for
evidence synthesis.

The broader narrative corpus also retained directed contextual and
methodological sources. These comprised foundational sources on general world
models and related technical methods; contextual sources on longitudinal
medical modelling, causal inference, medical foundation models, clinical
digital twins, mechanistic disease models, data resources, and relevant clinical domains; recent medical world-model studies that illustrate emerging application directions; and methodological sources
covering review methods, reporting guidance, fairness, privacy, uncertainty,
safety, evaluation, implementation, and governance. These sources were classified separately and were not
counted as strict empirical medical world-model studies. Assigning predefined
corpus roles allowed core empirical evidence to be distinguished from the
sources used for conceptual, methodological, and clinical interpretation.

\begin{figure}[!t]
\centering
\includegraphics[width=0.98\textwidth]{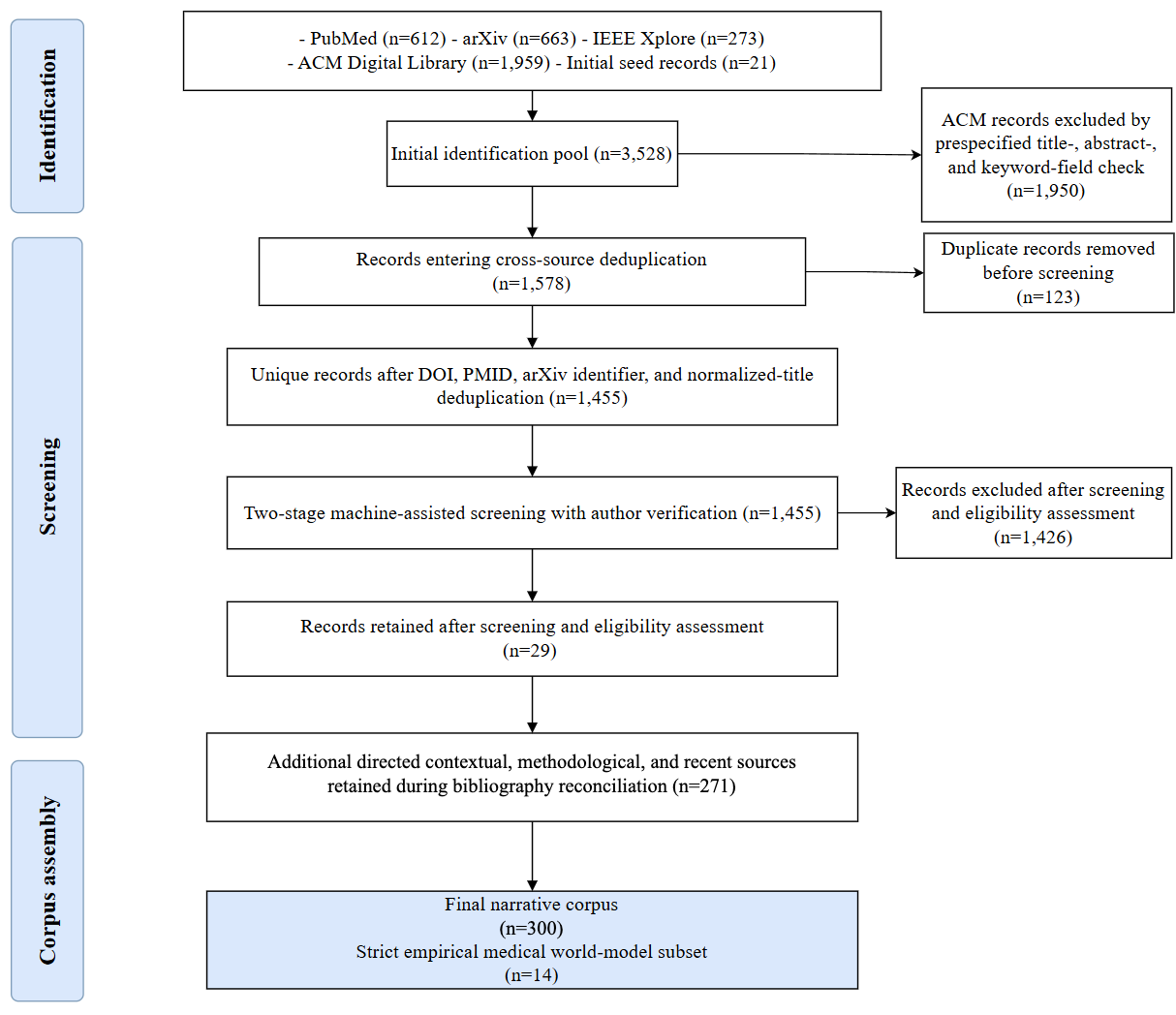}
\caption{Literature identification and selection flow. Electronic database searches and the initial seed set yielded 3,528 records.
Following prespecified ACM field filtering, cross-source deduplication, and
two-stage machine-assisted screening with author verification, 29 records
were retained after eligibility assessment. An additional 271 directed contextual,
foundational, methodological, and recent sources were retained through directed searches and
bibliography reconciliation, yielding a final narrative corpus of 300 sources,
including a strict empirical subset of 14 medical world-model studies.
This evidence-mapping flow is not presented as a PRISMA-ScR diagram.}
\label{fig:literature-selection-flow}
\end{figure}

\subsection{Record Processing and Study Selection}\label{sec:record-processing}

The four electronic database searches identified 3,507 records. Together
with the 21 initial seed records, 3,528 records entered the identification
pool. Because the ACM Digital Library search was intentionally broad, a
prespecified title-, abstract-, and keyword-field check was applied before
cross-source merging. This check excluded 1,950 of the 1,959 ACM records and
retained nine records satisfying the concept-field condition. After this ACM
prescreening, the 612 PubMed records, 663 arXiv records, 273 IEEE Xplore
records, nine ACM records, and 21 seed records yielded 1,578 records for
cross-source deduplication.

Duplicates were identified using DOI, PMID, arXiv identifier, and normalized
title. Crossref was used only to verify DOI and bibliographic metadata and to
help resolve relationships between preprints and published versions; it was
not counted as an independent identification source. Deduplication removed
123 records, leaving 1,455 unique records for screening.

Screening was conducted in two machine-assisted stages, with authors retaining
responsibility for the final eligibility decisions. The first stage combined
explicit eligibility rules with semantic relevance assessment to identify
both a medical context and a qualifying dynamic capability. Records with
incomplete metadata were retained for further assessment rather than excluded
automatically. The second stage assessed abstracts and available source
metadata, verified DOI and publication-version relationships where possible,
and assigned standardized disposition reasons, including exclusion for a
non-medical setting, absence of a qualifying dynamic model, or restriction to
a static task; classification as a background or review source rather than a
core empirical study; and consolidation as a duplicate publication. Authors
then reviewed the candidate set and reconciled eligible records with the manuscript
bibliography.

Following screening and author reconciliation, 29 database- or seed-derived
cited sources were retained. Through directed searches and bibliography reconciliation, 271
additional directed sources required for the
narrative synthesis were retained. They provided foundational, conceptual,
clinical, data-resource, review-methods, reporting, and trustworthy-AI support for the
narrative synthesis, together with recent medical world-model studies that illustrate emerging application directions. Bibliographic metadata for these sources were checked against Crossref or arXiv records. These additional sources were not treated as
electronically retrieved core empirical records and did not expand or alter the strict
empirical subset derived from the database and seed records.

The final manuscript-aligned narrative corpus therefore comprised 300 unique
cited sources ($29+271=300$). Of the 29 database- or seed-derived cited sources,
14 met the strict empirical definition of a medical world-model study. Thus,
the strict empirical subset was nested within the screened cited-source set,
rather than constituting a separate addition to the 300-source corpus. The
remaining sources supported conceptual, methodological, technical, or
clinical interpretation and were not represented as core empirical evidence.
Figure~\ref{fig:literature-selection-flow} summarizes the identification and
selection process.

\subsection{Data Charting and Evidence Synthesis}

For evidence-bearing medical studies, charted fields included publication status, medical domain, task, data modality and source, state representation, temporal scale or prediction horizon, action or intervention definition, dynamics architecture, rollout mechanism, L1--L4 capability level, evaluation design, clinician involvement, external or prospective validation, clinical endpoints, and availability of code and data. The synthesis used descriptive comparison and thematic mapping rather than meta-analysis because the included sources span heterogeneous tasks, modalities, architectures, and outcomes. Evidence was organized across six overlapping application tracks: medical-imaging representation and future-observation simulation; EHR and longitudinal patient-trajectory modelling; disease-progression and treatment-response simulation; physiological-signal and multimodal-state modelling; ultrasound guidance and surgical embodied interaction; and population- or health-system simulation.

\subsection{Evidence Appraisal}

Evidence maturity was described using publication status, external validation, prospective validation, clinician evaluation, clinical endpoints, and code or data availability. Technical capability and clinical evidence maturity were assessed separately. No single risk-of-bias score was imposed across the full corpus because it combines heterogeneous engineering studies, observational modelling studies, reviews, position papers, and reporting guidance. Limitations introduced by the machine-assisted, manuscript-aligned selection process are reported explicitly rather than interpreting the corpus as an exhaustive systematic sample. Capability levels for the recent studies summarized in Tables~\ref{tab:horizon_patient} and~\ref{tab:horizon_embodied} were assigned using the same L1--L4 definitions.

\section{Development History of Medical Artificial Intelligence}\label{sec:history}

\begin{figure}[!t]
\centering
\includegraphics[width=\textwidth]{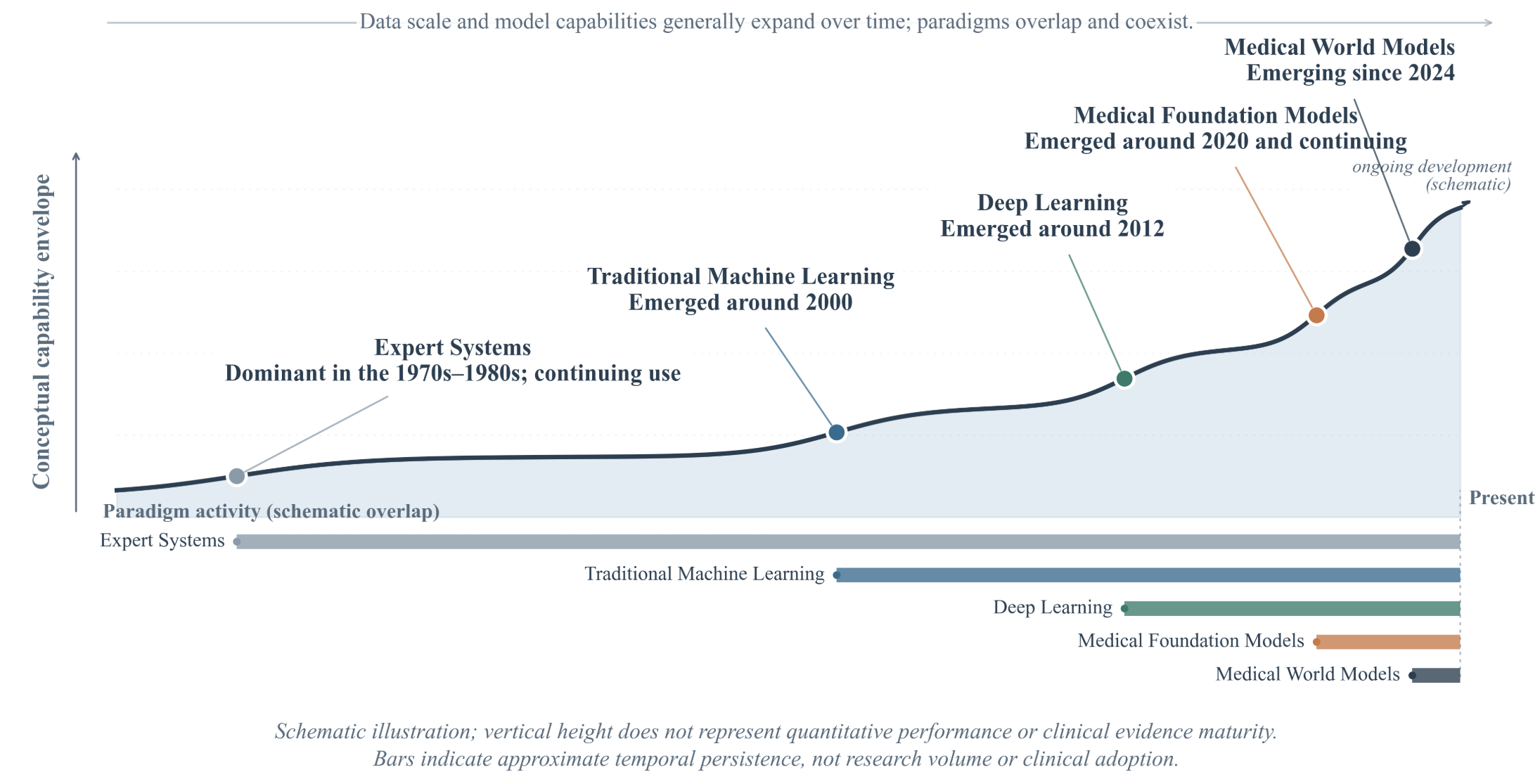}
\caption{Evolution of major paradigms in medical artificial intelligence. The figure traces the progression from expert systems and traditional machine learning to deep learning, medical foundation models, and emerging medical world models. The curve represents a qualitative expansion in data scale and modelling capability, whereas the horizontal bars indicate the approximate periods of continued use and temporal overlap among these paradigms. Neither the curve height nor the bar length represents quantitative performance, clinical evidence maturity, research volume, or clinical adoption.}\label{fig:medical_ai_evolution}
\end{figure}

Medical AI has evolved from knowledge-driven systems to data-driven methods, and then towards model-driven and world-model-driven paradigms. Across these stages, the research focus has expanded from disease recognition to disease prediction, treatment simulation, and clinical decision support. Core capabilities have continued to improve, while new limitations have also emerged. Overall, the development of medical AI can be divided into five stages: expert systems, traditional machine learning, deep learning, medical foundation models, and medical world models, as summarized in Table~\ref{tab:history}. Figure~\ref{fig:medical_ai_evolution} provides a visual summary of these five evolutionary stages and their capability leaps.

\begin{table}[!t]
\footnotesize
\setlength{\tabcolsep}{3pt}
\renewcommand{\arraystretch}{1.12}
\caption{Development stages of medical artificial intelligence. The table compares five overlapping stages, from expert systems to medical world models, in terms of core technologies, representative models or applications, capabilities, and limitations. The progression reflects a shift from rule-based diagnosis and feature-engineered prediction to representation learning, large-scale pretraining, and action-conditioned simulation. The periods indicate approximate times of emergence or dominant use, and greater technical capability does not necessarily imply stronger clinical validation or deployment readiness.}\label{tab:history}
\begin{tabularx}{\textwidth}{@{}YYYYY@{}}
\toprule
Development stage & Core technologies & Representative models/applications & Core capability & Main limitation\\
\midrule
Expert systems (1970s--1990s) & If--then rules; knowledge-base reasoning & MYCIN \cite{shortliffe1975mycin}; INTERNIST-1 \cite{miller1982internist} & Specialty-specific assisted diagnosis; interpretable reasoning & Difficult knowledge updating; limited generalization\\
Traditional machine learning (2000--2011) & SVM; random forest; logistic regression & Cardiovascular-risk models; pulmonary-nodule classifiers & Data-driven analysis; short-term risk prediction & Dependence on hand-crafted features; weak temporal modelling\\
Deep learning (2012--2019) & CNN/U-Net; RNN/Transformer & Medical image analysis \cite{gulshan2016retinopathy,esteva2017dermatologist}; ClinicalBERT \cite{ref003} as a precursor to medical foundation models & Automatic feature learning; task-specific prediction & Typically prediction-focused; lacks explicit, validated intervention-conditioned simulation\\
Medical foundation models (emerging since 2020) & Large-scale pretraining; multimodal unified modelling & Med-BERT \cite{ref007}; BEHRT \cite{ref106}; Foresight \cite{ref085}; MedSAM \cite{ref101}; UNI \cite{chen2024uni} & Cross-task transfer; few-shot generalization & Usually lacks explicit action-conditioned dynamics and clinically validated long-horizon planning\\
Medical world models (emerging since 2024) & Latent dynamics; action-conditioned simulation & Cardiac Copilot \cite{ref084}; EHRWorld \cite{ref047}; CLARITY \cite{ref070}; Brain-WM \cite{ref050} & Latent-state prediction; action-conditioned rollout in advanced systems; candidate counterfactual comparison & Limited clinical validation; limited causal identifiability\\
\botrule
\end{tabularx}
\end{table}

\subsection{Knowledge-driven expert systems}

Expert systems dominated the early development of medical artificial intelligence during the 1970s and 1980s and established the first generation of computer-assisted clinical decision support. Rooted in the symbolic-AI paradigm, these systems summarized clinical knowledge from medical experts and implemented automated reasoning through ``if--then'' rule bases. Representative systems included MYCIN and INTERNIST-1 \cite{shortliffe1975mycin,miller1982internist}, which could perform infectious-disease diagnosis, differential diagnosis, and treatment recommendation according to predefined rules, achieving reasoning performance close to clinical experts in specific specialties.

The main contribution of this paradigm was demonstrating that computers could emulate part of clinicians' diagnostic reasoning processes and assist medical decision-making. Nevertheless, because the knowledge base relied primarily on manually designed rules, updating clinical knowledge was labor-intensive, coverage of complex scenarios remained limited, and generalization to previously unseen diseases or therapeutic strategies was weak. Although rule-based systems continue to serve as important components in guideline engines and clinical decision-support systems today, they are no longer the dominant paradigm for medical AI research.

\subsection{Data-driven traditional machine learning}

With the widespread adoption of hospital information systems and electronic health records (EHRs) in the early 2000s, medical AI gradually shifted from knowledge-driven reasoning towards data-driven statistical learning. Support vector machines, random forests, logistic regression, and other traditional machine-learning algorithms became widely adopted for disease classification, risk prediction, and assisted diagnosis. Compared with expert systems, these methods learned statistical patterns directly from clinical data, reducing the dependence on manually encoded expert knowledge while improving model adaptability. Representative applications included cardiovascular-risk prediction, pulmonary-nodule malignancy classification, and ICU mortality-risk prediction.

Despite these advances, traditional machine-learning methods still depended heavily on manually engineered features. Their representational capacity was limited, making it difficult to model high-dimensional medical images, clinical text, and complex longitudinal patient records. Furthermore, most methods focused on single-time-point prediction and were unable to capture long-term disease progression or support dynamic treatment planning.

\subsection{Deep representation learning}

Beginning in 2012, breakthroughs in deep learning fundamentally transformed medical AI from handcrafted feature engineering to automatic representation learning. Convolutional neural networks (CNNs), recurrent neural networks (RNNs), Transformers, and U-Net enabled models to learn hierarchical semantic representations directly from medical images, pathology slides, and clinical text, substantially improving disease-recognition accuracy \cite{ronneberger2015unet,vaswani2017attention,gulshan2016retinopathy,esteva2017dermatologist}. Medical image analysis rapidly became one of the most successful application areas of deep learning. In parallel, Transformer-based architectures such as ClinicalBERT and BEHRT further advanced longitudinal EHR modelling and clinical outcome prediction by learning contextual representations from patient-care sequences \cite{ref003,ref106}. Recurrent models such as Doctor AI, and deep-learning pipelines applied to raw EHR data from two academic medical centres, showed that longitudinal care sequences could be used to predict future clinical events and outcomes \cite{choi2016doctorai,rajkomar2018scalable}.

Although deep learning significantly improved predictive performance, most models remained discriminative systems that learned mappings from observations to outputs. They lacked explicit modelling of disease dynamics and intervention-dependent state transitions, making it difficult to answer counterfactual clinical questions such as ``what would happen if an alternative treatment strategy were adopted?'' Consequently, deep learning primarily advanced perception intelligence rather than long-term clinical decision intelligence.

\subsection{Medical foundation models}

Since around 2020, large-scale self-supervised pretraining has driven medical AI into the era of foundation models \cite{bommasani2021opportunities,moor2023foundation}. Medical foundation models are pretrained on massive multimodal medical datasets to learn transferable representations that can be shared across multiple downstream tasks. Recent surveys have systematically summarized the development of medical foundation models and their future opportunities \cite{ref026,ref031}. In the EHR domain, Med-BERT, BEHRT, ClinicalBERT, Foresight, and TransformEHR have substantially improved disease prediction and longitudinal patient modelling \cite{ref007,ref106,ref003,ref085,ref100}. In medical imaging, MedSAM and RETFound have demonstrated strong cross-organ and cross-modal generalization capabilities \cite{ref101,ref103}, and pathology foundation models such as UNI and Prov-GigaPath have extended this paradigm to whole-slide images \cite{chen2024uni,xu2024gigapath}. In addition, multimodal generative models and medical large language models have significantly enhanced medical knowledge understanding, report generation, and clinical question answering \cite{ref102,singhal2025medpalm2,jiang2023nyutron}.

Compared with previous paradigms, medical foundation models greatly alleviate annotation requirements while enabling few-shot learning, transfer learning, and cross-task generalization. However, their primary objective remains learning statistical associations among multimodal observations. Although these models can predict future outcomes, they generally do not explicitly model latent patient states, disease dynamics, or intervention-conditioned state transitions. Consequently, they remain limited in supporting long-horizon disease simulation, treatment-response modelling, and clinically validated decision planning \cite{ref056,ref095}.

A notable intermediate development is the emergence of generative models of entire health trajectories. Foresight, ETHOS, CoMET, and Delphi-2M are trained autoregressively on tokenized health events and can sample plausible future timelines rather than single endpoints \cite{ref085,renc2024ethos,ref041,shmatko2025delphi}, and conditional generative models have simulated detailed multivariable trajectories of Alzheimer's disease progression \cite{fisher2019adforecast}. These models move beyond static association towards temporal simulation, but they generally represent treatments as tokens in the same stream as outcomes, so that intervention semantics remain implicit and the learned dynamics mix disease biology with care processes. Recent work accordingly argues that patients should be modelled as dynamical systems whose states evolve under interventions rather than as documents to be continued token by token \cite{adam2026smbstructure}. This distinction marks the conceptual boundary between generative foundation models and medical world models.

\subsection{Medical world models}

More recently, medical world models have emerged as a promising paradigm for moving medical AI beyond static prediction towards dynamic simulation and decision support. Instead of treating patients as collections of independent observations, medical world models regard patients as continuously evolving dynamic systems. By jointly modelling latent patient states, disease dynamics, and clinical interventions, they aim to learn how patient states evolve over time and thereby support disease-course prediction, treatment-response simulation, counterfactual reasoning, and long-term clinical planning \cite{ref056,ref095}.

Representative studies have rapidly expanded across multiple clinical domains. EHRWorld proposes long-horizon clinical-trajectory simulation; EchoJEPA learns predictive latent representations from echocardiographic videos; CLARITY and MeWM model treatment-conditioned disease evolution; Brain-WM jointly predicts treatment sequences and future glioblastoma imaging; and EchoWorld, Cardiac Copilot, and SAW extend world-model concepts to ultrasound guidance and surgical video generation \cite{ref047,ref046,ref070,ref039,ref050,ref074,ref084,ref051}. ChronoMedicalWorld further models action-conditioned chronic-disease trajectories using structured interventions and patient--care communication \cite{ref054}. By contrast, SteeraMed Core is better interpreted as an emerging mechanism-alignment evidence-chain framework: it has demonstrated retrospective positive-control recovery but has not yet implemented a calibrated state--action--next-state learning system or established clinical efficacy \cite{ref125}.

Overall, medical world models represent an important conceptual shift from recognizing patient states to understanding disease evolution and ultimately simulating future clinical trajectories under different interventions. Nevertheless, this field remains in its early stages. Most existing systems are still developed using retrospective datasets, while challenges including causal inference, trustworthy evaluation, uncertainty quantification, model safety, and regulatory compliance remain major obstacles before large-scale clinical deployment becomes feasible \cite{ref013,ref098,ref099,ref044,ref053,ref064,ref066,ref067,ref089,ref090,ref092,ref097,ref104,ref105,ref107,ref108,ref109,ref114,ref123,ref124}.

\FloatBarrier

\section{Core Concepts and Technical Framework of Medical World Models}
\label{sec:framework}

Medical world models extend medical AI beyond direct outcome prediction by
representing patients as evolving dynamic systems and modelling how their
states change over time and in response to clinical interventions. Whereas
many conventional prediction models estimate a diagnosis, risk, or future
outcome directly from observed data, medical world models introduce an
internal patient-state representation together with transition dynamics that
can support temporal simulation under specified clinical actions. This
framework may enable disease-course forecasting, treatment-response
simulation, candidate-intervention comparison, and planning-oriented clinical
decision support \cite{ref056,ref095}. These functional capabilities should,
however, be distinguished from evidence of causal validity or clinical
effectiveness.

A conventional longitudinal prediction model can be represented by the
conditional distribution
\begin{equation}
P(y_{t+1}\mid x_{1:t}),
\label{eq:traditional_prediction}
\end{equation}
where \(x_{1:t}=(x_1,\ldots,x_t)\) denotes the sequence of observations
available up to time \(t\), and \(y_{t+1}\) denotes the outcome to be
predicted at the subsequent time point. This formulation can capture
temporal associations between historical observations and future clinical
outcomes. By itself, however, it does not specify an explicit latent-state
transition process that can be recursively simulated, nor does it describe
how future trajectories may change under alternative clinical actions.

In contrast, medical world models introduce latent patient states and
state-transition dynamics to represent how clinical conditions evolve over
time and, in more advanced systems, in response to interventions. Rather
than predicting only an endpoint from the observed history, they model the
transition from a current patient state to a subsequent state, potentially
conditioned on treatment, procedure, acquisition, or device-control actions.
This formulation provides the computational basis for multi-step trajectory
simulation, treatment-response modelling, and candidate-intervention
comparison \cite{ref056,ref095,ref047}. Nevertheless, action-conditioned
prediction alone should not be interpreted as valid counterfactual inference;
such claims additionally require well-defined interventions, defensible
causal assumptions, uncertainty estimation, and appropriate clinical
validation.
The mathematical formulation of patient-state transitions is presented in
Section~\ref{subsec:dynamics}.

\begin{figure}[!t]
\centering
\includegraphics[width=0.98\textwidth]{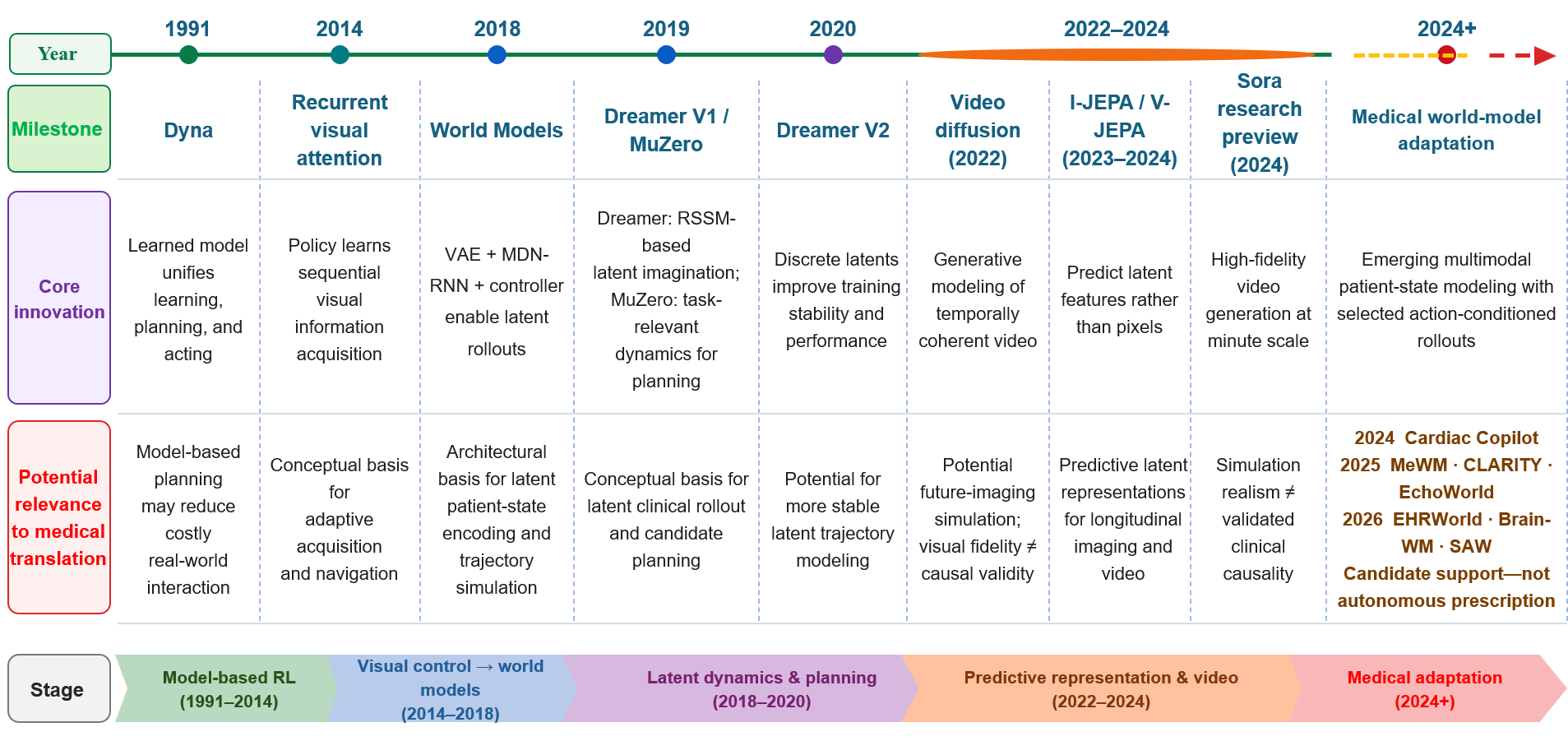}
\caption{Technical evolution of world models and their adaptation to medicine. The timeline traces selected milestones from early model-based reinforcement learning to latent-dynamics modelling, predictive representation learning, diffusion and video-generation approaches, and emerging medical world models. The milestones illustrate developments in learned dynamics, internal rollout, future-state prediction, and action-conditioned simulation that have informed medical applications. Years indicate the first public release of the corresponding work, including a preprint or public preview where applicable. The timeline is selective rather than exhaustive and does not represent comparative model performance, clinical evidence maturity, or readiness for clinical deployment.}\label{fig:technical_timeline}
\end{figure}

\begin{figure}[!t]
\centering
\includegraphics[width=0.98\textwidth]{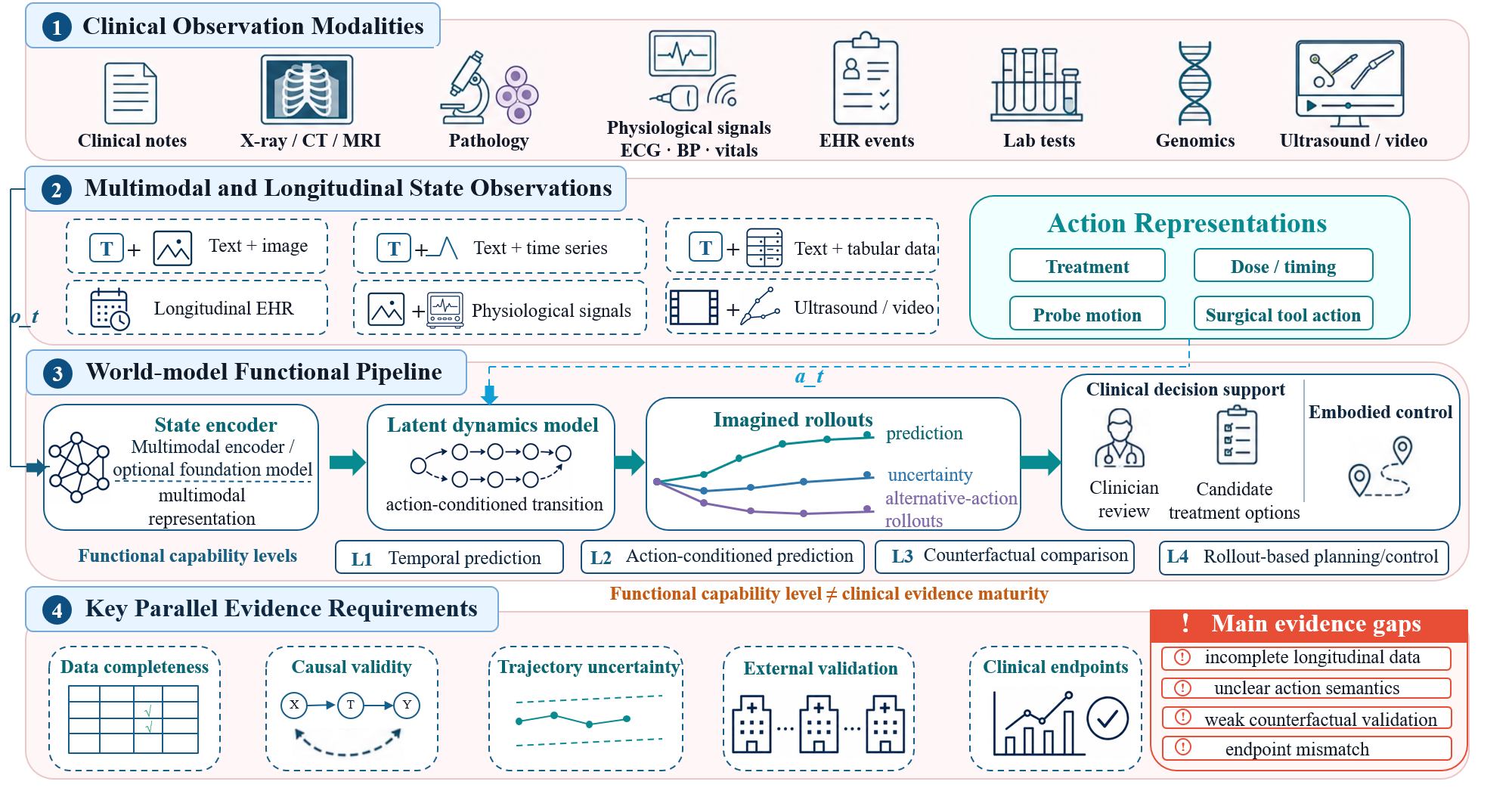}
\caption{Overview of medical world models in medicine. Multimodal and
longitudinal clinical observations are encoded into patient-state
representations and combined with explicit action representations,
latent-dynamics models, imagined rollouts, and planning, decision-support,
or embodied-control modules. The L1--L4 levels describe functional
capability rather than clinical evidence maturity. Trustworthy translation
requires parallel evidence for data completeness, causal validity,
trajectory-level uncertainty, external validation, and clinically meaningful
endpoints, while current studies remain limited by incomplete longitudinal
data, unclear action semantics, weak counterfactual validation, and endpoint
mismatch. Decision-support outputs should therefore be interpreted as
candidate options for clinician review rather than autonomous
prescriptions.}\label{fig:mwm_overview}
\end{figure}

\begin{figure}[!t]
\centering
\includegraphics[width=0.98\textwidth]
{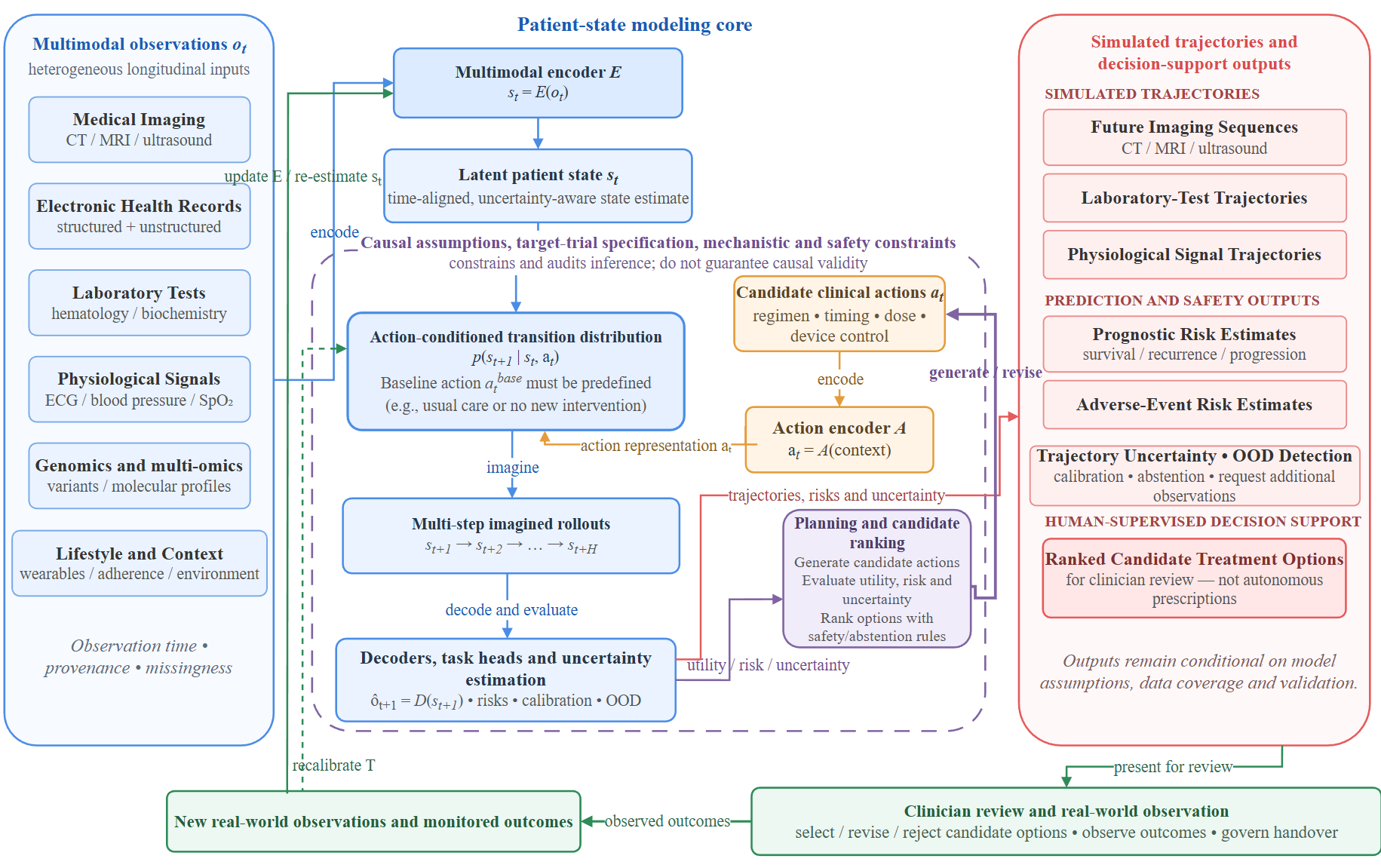}
\caption{Conceptual architecture of medical world models. Multimodal
clinical observations are encoded into latent patient states, whose
evolution is modelled through temporal and intervention-conditioned
transitions. Multi-step rollouts may generate future imaging observations,
laboratory and physiological trajectories, prognostic and adverse-event
risks, and candidate treatment strategies. Causal or mechanistic
constraints, uncertainty estimation, safety constraints, and candidate-plan
ranking are presented as design requirements rather than established
clinical capabilities. New observations update the estimated patient state
and recalibrate the model, while all decision-support outputs remain subject
to clinician review.}
\label{fig:mwm_architecture}
\end{figure}

Figure~\ref{fig:technical_timeline} situates this paradigm shift within the
broader technical evolution of world models. Dyna established an early
framework that integrated learning, planning, and acting through a learned
environment model, whereas recurrent visual-attention models demonstrated
sequential decision-making through reinforcement learning
\cite{sutton1991dyna,mnih2014recurrent}. Building on these foundations,
medical world models adapt world-model principles to clinical settings by
jointly representing multimodal patient observations, patient-state dynamics,
and intervention-conditioned transitions. Figure~\ref{fig:mwm_overview}
summarizes the conceptual organization of medical world models, and
Figure~\ref{fig:mwm_architecture} illustrates their closed-loop architecture,
linking multimodal observations, latent patient-state representation,
disease-dynamics modelling, clinical actions, simulated trajectories, and
clinician-supervised decision support.

\subsection{Medical world models and adjacent paradigms}\label{subsec:adjacent_paradigms}

Because the term ``world model'' is applied to heterogeneous systems, we use four operational criteria to delineate medical world models from adjacent paradigms: (i) a representation of an evolving state that is only partially observed; (ii) a learned transition mechanism that can be applied recursively; (iii) at higher capability levels, the conditioning of transitions on specified actions; and (iv) the use of internal rollouts for forecasting, comparison, or planning. Table~\ref{tab:concepts} contrasts medical world models with clinical prediction models, medical foundation models, generative models, causal treatment-effect models, mechanistic simulators, clinical digital twins, and reinforcement-learning policies. These categories overlap in practice: a foundation model may serve as the encoder of a world model, a mechanistic simulator may constrain its transitions, a causal estimand may define what its rollouts should approximate, and a digital twin may embed a world model as its predictive core. The distinction nevertheless matters, because each category carries a different implicit claim and therefore requires different evidence.

\begin{table}[!t]
\scriptsize
\setlength{\tabcolsep}{2pt}
\renewcommand{\arraystretch}{1.08}
\caption{Medical world models and adjacent paradigms. The columns indicate whether each paradigm typically maintains an explicit state, conditions on explicit actions, and performs recursive rollout, together with its typical form of validation and its relation to medical world models. Entries describe typical rather than universal practice, and the categories frequently overlap.}\label{tab:concepts}
\begin{tabularx}{\textwidth}{@{}>{\raggedright\arraybackslash}p{0.13\textwidth}>{\raggedright\arraybackslash}p{0.19\textwidth}>{\raggedright\arraybackslash}p{0.08\textwidth}>{\raggedright\arraybackslash}p{0.08\textwidth}>{\raggedright\arraybackslash}p{0.08\textwidth}YY@{}}
\toprule
Paradigm & Central question & Explicit state & Explicit action & Recursive rollout & Typical validation & Relation to medical world models\\
\midrule
Clinical prediction model & What is the risk of an outcome at a given horizon? & Rarely & Rarely (as covariate) & No & Discrimination, calibration, external validation \cite{ref066,vancalster2019calibration} & Supplies endpoints, baselines, and reporting standards\\
Medical foundation model & Which transferable representation supports many tasks? & Implicit embedding & No & No & Downstream transfer and few-shot performance \cite{moor2023foundation,ref026} & Candidate encoder of the patient state\\
Generative model & Which observations are plausible? & Implicit & Via conditioning & Sometimes (autoregressive) & Fidelity, diversity, expert realism \cite{kazerouni2023diffusion,ref014} & Decoder or simulator component; realism does not establish validity\\
Causal treatment-effect model & What is the effect of one intervention versus another? & Summary covariates & Yes (treatment) & Rarely & Identification assumptions, sensitivity analysis, trial benchmarking \cite{feuerriegel2024causal,hernan2016target} & Supplies estimands and identification conditions for counterfactual claims\\
Mechanistic or physiological simulator & How does the system evolve under known laws? & Yes (physiological parameters) & Yes & Yes & Parameter calibration; verification, validation, and uncertainty quantification \cite{man2014uvapadova,nasem2024digitaltwins} & Constraint, prior, or hybrid component of transitions\\
Clinical digital twin & How will this individual's synchronized virtual counterpart evolve and respond? & Yes (patient-specific) & Yes & Yes, with continual updating & Patient-specific calibration and synchronization \cite{ref110,venkatesh2022healthdt,corral2020digitaltwin} & Integration framework that may embed a world model\\
Reinforcement-learning policy & Which action should be taken now? & Yes (decision state) & Yes & Only if model-based & Off-policy evaluation, simulation \cite{gottesman2019guidelines,komorowski2018aiclinician} & Consumer of world-model rollouts in planning\\
Medical world model & How will the patient, procedural, or population state evolve, and how would it evolve under alternative actions? & Yes (latent) & At L2--L4 & Yes & Capability-specific (Table~\ref{tab:claims_evidence}) & ---\\
\botrule
\end{tabularx}
\end{table}

\subsection{Patient-state representation: from multimodal observations to a unified latent space}
\label{subsec:state_representation}

Patient-state representation provides the foundation for temporal prediction
and internal rollout in medical world models. In clinical practice, a
patient's underlying health state is only partially observable and must be
inferred from heterogeneous data, including medical images, electronic
health records (EHRs), physiological signals, laboratory measurements, and
multi-omics profiles \cite{ref047,ref110,ref117}. These data sources differ
substantially in temporal resolution, sampling frequency, scale, and
measurement uncertainty. Medical world models therefore require a unified
representation that integrates complementary observations while preserving
clinically relevant temporal and modality-specific information.

In a simplified formulation, a multimodal encoder maps the observations
available at time \(t\) to a latent patient state:
\begin{equation}
s_t = E_{\psi}(o_t),
\label{eq:encoder}
\end{equation}
where \(o_t\) denotes the set of multimodal clinical observations available
at time \(t\), \(E_{\psi}\) denotes an encoder parameterized by \(\psi\), and
\(s_t\) is the corresponding latent patient-state representation. When the
current state is inferred from longitudinal history rather than
contemporaneous observations alone, the encoder may instead operate on
\(o_{1:t}\). Eq.~(\ref{eq:encoder}) is therefore intended as a general
representational abstraction rather than a specification of a particular
encoder architecture.

The use of compact latent states can be traced to the variational
autoencoder-based architecture of World Models \cite{ref082}. Dreamer
subsequently introduced recurrent state-space models (RSSMs) that combine
deterministic and stochastic latent components, enabling recurrent state
estimation and imagined rollout \cite{ref002,ref005,ref016,ref008}.
Joint-embedding predictive architectures (JEPAs) further shifted the emphasis
from reconstructing raw observations to predicting future representations,
thereby encouraging the learning of temporally informative latent features
\cite{ref017}.

In medical settings, an informative patient state should capture more than
disease category or stage. It may also encode organ function, physiological
stability, treatment history and response, comorbidities, and
patient-specific variation. EHRWorld and ChronoMedicalWorld infer evolving
clinical states from longitudinal care records, whereas EchoJEPA learns
predictive representations from echocardiographic video
\cite{ref047,ref054,ref046}. These examples illustrate that patient-state
representations are modality- and task-dependent; no single latent
architecture has yet emerged as a standard across medical world models.

Clinical observations are also irregularly sampled and informatively missing. Laboratory tests are ordered because clinicians suspect a change, so the presence and timing of a measurement can carry more prognostic information than its value: across 272 common laboratory tests in two hospitals, the presence of a test order was significantly associated with survival for 86\% of tests \cite{agniel2018biases}. State encoders therefore need to model the observation process itself. Available mechanisms include decay-based recurrent units such as GRU-D, and continuous-time latent ordinary differential equations and neural controlled differential equations that update the state at arbitrary observation times \cite{che2018grud,chen2018neuralode,rubanova2019latentode,kidger2020ncde}. Multimodal fusion adds a further layer of complexity, because imaging, text, physiological signals, and omics differ in resolution, semantics, and missingness \cite{acosta2022multimodal}. Pretrained encoders such as RETFound, UNI, and Prov-GigaPath can provide strong modality-specific features \cite{ref103,chen2024uni,xu2024gigapath}, but embeddings optimized for static recognition do not necessarily preserve the variables that govern disease dynamics. Recent EHR studies accordingly train encoders to predict future patient states in latent space rather than only the next token, so that trajectory dynamics must be encoded before the next state is observed \cite{adam2026smbstructure,yang2026clinjepa}.

\subsection{Disease-dynamics modelling: learning longitudinal patient evolution}
\label{subsec:dynamics}

The dynamics model characterizes how a latent patient state evolves over
time and distinguishes medical world models from direct outcome predictors.
It defines a transition distribution between successive states, potentially
conditioned on a clinically relevant action:

\vspace{-1cm}
\begin{equation}
P(s_{t+1}\mid s_t,a_t),
\label{eq:dynamics}
\end{equation}
\vspace{-1cm}

\noindent where \(s_t\) and \(s_{t+1}\) denote the latent patient states at
consecutive time points, and \(a_t\) represents a treatment, procedure,
acquisition action, device-control command, or other relevant clinical
condition at time \(t\).

When actions are not explicitly modelled, the transition reduces to
\(P(s_{t+1}\mid s_t)\) and describes state evolution under the care
conditions represented in the observed data. These trajectories should not
be interpreted as untreated disease progression, because they may already
reflect routine care, treatment selection, adherence, and other time-varying
influences. When actions are incorporated, the model estimates how the
distribution of future states varies under specified interventions, thereby
supporting treatment-response simulation and action-conditioned rollout.
Such predictions remain associational unless the analysis also specifies
the intervention, addresses confounding, and states the causal assumptions
required for counterfactual interpretation.

Existing dynamics models can be grouped into three broad, non-mutually
exclusive classes. Recurrent state-space models, exemplified by the
RSSM-based Dreamer family, support efficient latent-state estimation and
imagined rollout, but remain susceptible to model bias and error
accumulation over long horizons \cite{ref002,ref005,ref016}.
Transformer-based and recurrent latent-transition models are well suited to
capturing long-range dependencies in longitudinal EHRs and clinical-event
sequences \cite{ref085,ref003,ref106,ref100,ref054}. Diffusion- and
flow-based models instead learn distributions over future observations and
have been applied to tumour evolution and longitudinal brain MRI. However,
the visual plausibility of generated trajectories does not, by itself,
establish clinically valid treatment effects
\cite{ref039,ref014,ref020,ref050}.

These learned approaches inherit a longer tradition of disease-progression modelling. Event-based models and Subtype and Stage Inference reconstruct the ordering of biomarker abnormalities and identify disease subtypes from largely cross-sectional data \cite{fonteijn2012ebm,young2018sustain}; deep Markov models combine neural emission and transition functions with structured variational inference \cite{krishnan2017dmm}; and conditional generative models have simulated individual trajectories of dozens of clinical variables in Alzheimer's disease \cite{fisher2019adforecast}. In imaging, sequence-aware and latent diffusion models generate future medical images conditioned on earlier scans and covariates \cite{yoon2023sadm,puglisi2024brlp}, building on denoising diffusion, latent diffusion, and flow-matching formulations \cite{ho2020ddpm,rombach2022ldm,lipman2023flow}.

A second line of work embeds mechanistic knowledge directly into the transition model. Physics-informed neural networks penalize violations of governing equations \cite{raissi2019pinn}; hybrid models integrate expert pharmacological ordinary differential equations with neural latent dynamics \cite{qian2021expertode}; and neural-ODE pharmacokinetic models have shown advantages over alternative machine-learning models when predicting responses to new dosing regimens \cite{lu2021neuralodepk}. In oncology, reaction--diffusion and classical growth laws provide interpretable priors for tumour evolution \cite{swanson2000glioma,benzekry2014tumor}, and image-calibrated mechanistic models have forecast spatially resolved responses of high-grade glioma to chemoradiation in a small patient cohort \cite{hormuth2021glioma,wu2022oncologytwin}. Such constraints can reduce physiologically implausible rollouts, but they also import the assumptions and parameter-identifiability limits of the mechanistic model, which must be validated in their own right.

Medical dynamics modelling is further complicated by partial observability.
Patient trajectories are influenced by age, genetic background,
comorbidities, lifestyle, treatment adherence, access to care, and other
factors that are often incompletely recorded. Clinical observations also
differ in sampling frequency, measurement uncertainty, and missingness
mechanisms. Medical world models should therefore represent a distribution
of plausible future trajectories and quantify uncertainty, rather than
return only a single average prediction \cite{ref065,ref097}.

\subsection{Action-conditioned simulation and counterfactual intervention analysis}
\label{subsec:action_counterfactual}

Action conditioning extends medical world models beyond passive trajectory
prediction by explicitly representing the relation between clinical actions
and subsequent patient states. Conventional prediction models may include
treatment history as an input, but the formulation in
Eq.~(\ref{eq:traditional_prediction}) does not, by itself, specify a
transition process that can be repeatedly applied to simulate alternative
future action sequences.

Medical world models instead use the transition formulation in
Eq.~(\ref{eq:dynamics}) to condition predicted state evolution on clinically
relevant actions, including medication, radiotherapy, surgery, mechanical
ventilation, and treatment timing. Applying the model under different
candidate action sequences enables comparison of predicted patient
trajectories and provides a computational basis for treatment-response
modelling and planning-oriented decision support
\cite{ref056,ref095}.
Intervention-conditioned simulation should nevertheless be distinguished
from causal counterfactual inference. Differences between predicted
trajectories may reflect treatment-selection bias, time-varying confounding,
incomplete action documentation, or model misspecification rather than the
causal effects of treatment. Counterfactual interpretation therefore
requires clearly defined interventions and outcomes, an explicit causal
estimand, defensible assumptions regarding treatment assignment and
confounding, and validation against appropriate clinical evidence.
Relevant methodological foundations include potential-outcome frameworks,
causal Transformers, target-trial emulation, and individualized
treatment-effect estimation
\cite{ref013,ref065,ref006,ref012,ref075,ref098,ref099}.

The causal-inference literature offers several method families that can either be embedded in medical world models or used to audit them (Table~\ref{tab:causal_methods}). The potential-outcomes framework defines effects as contrasts between outcomes under hypothetical interventions \cite{rubin1974potential}. For time-varying treatments, Robins' g-formula expresses the distribution of outcomes under a treatment strategy as a sequence of conditional covariate and outcome models, a structure that closely parallels a world-model rollout \cite{robins1986gformula}; marginal structural models instead reweight observed trajectories by inverse probabilities of treatment \cite{robins2000msm}; and dynamic treatment regimes formalize sequential decision rules that adapt to evolving patient states \cite{murphy2003dtr}. Deep-learning implementations include G-Net for g-computation with recurrent networks, adversarially balanced and Transformer-based representations for counterfactual outcomes over time, and neural controlled differential equations for continuous-time treatment effects \cite{li2020gnet,ref006,ref012,seedat2022tecde}. The Time Series Deconfounder addresses multi-cause hidden confounding under explicit structural assumptions \cite{bica2020deconfounder}, and target-trial emulation provides a design framework for specifying eligibility, treatment strategies, time zero, and follow-up before modelling begins \cite{hernan2016target}. Under consistency, sequential exchangeability, and positivity, the expected outcome under a treatment strategy \(\bar a=(a_0,\ldots,a_K)\) is identified by the g-formula
\begin{equation}
\mathbb{E}\!\left[Y^{\bar a}\right]=\int \mathbb{E}\!\left[Y\mid \bar L_K=\bar l_K,\bar A_K=\bar a\right]\prod_{k=0}^{K}p\!\left(l_k\mid \bar l_{k-1},\bar a_{k-1}\right)d\bar l_K,
\label{eq:gformula}
\end{equation}
where \(\bar L_k\) denotes the history of time-varying covariates up to time \(k\), \(\bar A_k\) the corresponding treatment history, and \(Y^{\bar a}\) the potential outcome under strategy \(\bar a\) \cite{robins1986gformula}. A world model that recursively samples \(l_{k+1}\) given the simulated history and the queried actions implements a Monte Carlo approximation of Eq.~(\ref{eq:gformula}). Conversely, when exchangeability or positivity fails, or when the transition models are misspecified in regions visited by the rollout, the same procedure reproduces historical treatment-selection patterns rather than interventional effects. CaMeWM makes this connection explicit by restricting queried regimens to histories with empirical treatment support and by evaluating its induced policy with cross-fitted doubly robust estimation \cite{gao2026camewm}.

\begin{table}[!t]
\scriptsize
\setlength{\tabcolsep}{2pt}
\renewcommand{\arraystretch}{1.08}
\caption{Causal method families relevant to medical world models. Each family specifies how effects of time-varying interventions can be defined, identified, or estimated, and indicates how it may be embedded in or used to audit action-conditioned rollouts. No method removes the need for well-defined interventions and defensible identification assumptions.}\label{tab:causal_methods}
\begin{tabularx}{\textwidth}{@{}>{\raggedright\arraybackslash}p{0.15\textwidth}YYY>{\raggedright\arraybackslash}p{0.13\textwidth}@{}}
\toprule
Method family & Core idea & Key assumptions & Role in medical world models & Representative sources\\
\midrule
Potential outcomes and target-trial emulation & Define estimands as contrasts of outcomes under hypothetical interventions; emulate the protocol of a randomized trial with observational data & Consistency, exchangeability, positivity; well-defined time zero and strategies & Specifies the question that a rollout must answer before any modelling & \cite{rubin1974potential,hernan2016target,ref098}\\
g-formula and g-computation & Model time-varying covariates and outcomes sequentially, then simulate under the queried strategy & Sequential exchangeability; correctly specified transition and outcome models & Structurally analogous to action-conditioned rollout; errors compound with horizon & \cite{robins1986gformula,li2020gnet}\\
Marginal structural models and inverse weighting & Reweight observed trajectories by inverse probabilities of treatment & Correct treatment model; positivity & Audits rollouts and supports doubly robust evaluation of model-derived policies & \cite{robins2000msm,gao2026camewm}\\
Balanced sequence representations & Learn history representations that predict outcomes while reducing treatment-assignment bias & Sequential ignorability; adequate overlap & Transition encoders with reduced confounding bias & \cite{ref006,ref012}\\
Continuous-time treatment-effect models & Neural differential equations for irregularly sampled covariates and treatments & As above, plus an adequate observation model & Align counterfactual rollout with irregular clinical measurement & \cite{seedat2022tecde,kidger2020ncde}\\
Hidden-confounding and sensitivity methods & Infer substitute confounders from multiple treatments; bound bias with E-values or negative controls & Multi-cause structure; plausible sensitivity parameters & Quantify the robustness of counterfactual rollouts to unmeasured confounding & \cite{bica2020deconfounder,vanderweele2017evalue,lipsitch2010negative}\\
Dynamic treatment regimes and off-policy evaluation & Estimate the value of adaptive decision rules from observational trajectories & Positivity for the evaluated policy; correct reward specification & Required before interpreting planning outputs as improved care & \cite{murphy2003dtr,thomas2016ope,gottesman2018evaluating}\\
\botrule
\end{tabularx}
\end{table}

CLARITY, Brain-WM, Medical World Model for tumour evolution, and
world-model-enhanced offline reinforcement learning provide early examples
of treatment-conditioned trajectory modelling and candidate-strategy
evaluation \cite{ref070,ref050,ref039,ref049}. These studies illustrate the
potential of medical world models for individualized treatment analysis,
although current evidence remains largely retrospective or
simulation-based. Their outputs should therefore be interpreted as
model-dependent comparisons for clinical review rather than validated
treatment recommendations.

\begin{table}[!t]
\footnotesize
\setlength{\tabcolsep}{1.5pt}
\renewcommand{\arraystretch}{1.03}
\caption{Core system components and implementation strategies for medical world models. Six layers are considered: patient state representation, transition modelling, action representation, causal and mechanistic constraints, planning, and embodied interaction. Representative approaches are compared according to their functional roles, potential advantages, and limitations in medical settings. Foundation models and predictive representation methods can support patient state representation, whereas medical world models additionally require an explicit account of temporal state evolution. Rollout, action conditioning, and interaction mechanisms enable more advanced simulation and planning capabilities.}\label{tab:technical_routes}
\begin{tabularx}{\textwidth}{@{}YYYYY@{}}
\toprule
System layer & Components or routes & Primary role & Main advantages & Limitations and medical context\\
\midrule
State representation & Multimodal encoders; foundation-model-assisted encoders; JEPA/V-JEPA & Encode multimodal observations into unified latent patient states; predictive variants learn temporally informative representations & Cross-modal integration; representation transfer; reduced reconstruction burden & Not a dynamics or planning system alone; modality alignment and missingness remain difficult\\
Transition modelling & RSSM/recurrent state-space; Transformer/recurrent transition; flow/diffusion dynamics & Predict future latent states or observations & Efficient latent rollout and long-range dependency modelling; detailed generation when needed & State-estimation and rollout errors may accumulate; fine detail may be lost or generation may be costly\\
Action representation & Structured action ontologies; language-conditioned actions; device trajectories & Specify treatment, acquisition, or control conditions & Supports heterogeneous clinical and physical actions & Ambiguous granularity, incomplete documentation, and costly annotation\\
Causal design and mechanistic constraints & Target-trial alignment; structural causal models; pharmacokinetic or physical constraints & Define clinically meaningful interventions and estimands, and constrain implausible state transitions & Clarifies causal assumptions and improves mechanistic plausibility & Unmeasured confounding and incomplete mechanisms remain unresolved\\
Planning & MBRL; model predictive control; candidate-plan ranking & Evaluate rollouts and rank candidate actions & Potentially more sample-efficient than model-free RL; supports constrained planning & Model bias, offline distribution shift, unobserved confounding, and safety constraints\\
Embodied interaction & Probe guidance; surgical-tool control; robotic navigation & Close the perception--simulation--control loop & Handles continuous actions and physical interaction & Hardware variation, domain shift, real-time control, and safety validation\\
\botrule
\end{tabularx}
\vspace{2pt}\par\scriptsize\textit{Abbreviations:} JEPA, Joint-Embedding Predictive Architecture;
V-JEPA, Video Joint-Embedding Predictive Architecture;
RSSM, Recurrent State-Space Model;
MBRL, model-based reinforcement learning;
RL, reinforcement learning.
\end{table}

\subsection{Clinical planning and decision-support optimization}\label{subsec:planning}

The ultimate goal of a medical world model is not only to predict the future, but to support candidate long-term clinical planning inside an internal simulated environment. Traditional reinforcement learning relies on repeated trial and error in the real environment, which is infeasible in medicine. A world model first builds a digital patient environment and then performs multi-step rollout within that simulated space \cite{ref082,ref002,ref005,ref004,ref016}.

Planning involves generating plausible future trajectories, comparing the
expected benefits and risks of candidate interventions, and ranking
strategies against predefined clinical objectives. This approach is being
explored in individualized treatment planning and healthcare-resource
allocation. 
CLARITY, MeWM, and Policy4OOD provide early examples of
candidate-strategy evaluation, although retrospective or simulation-based
results do not establish prospective clinical benefit
\cite{ref070,ref039,ref048}. 
SteeraMed Core instead offers a
mechanism-alignment framework and a pathway towards transition modelling,
rather than a validated rollout or planning system \cite{ref125}.

Planning within a learned model has a characteristic failure mode: an optimizer tends to seek out regions in which the model is wrong but optimistic. Model-based reinforcement learning therefore pairs dynamics learning with uncertainty-aware planning, for example by sampling trajectories from probabilistic ensembles \cite{chua2018pets}, and offline model-based methods penalize rollouts that leave the support of the data \cite{yu2020mopo,kidambi2020morel}. Conservative value estimation offers an alternative for purely offline settings \cite{kumar2020cql}, and reviews of offline reinforcement learning identify distribution shift between the behaviour policy and the learned policy as the central obstacle \cite{levine2020offline}. Scalable latent planners such as TD-MPC2 combine learned dynamics with model-predictive control and have recently been transferred to endovascular navigation \cite{hansen2024tdmpc2,robertshaw2025mtwm,robertshaw2026safeendo}.

Experience with reinforcement learning in healthcare illustrates the stakes. The AI Clinician learned fluid and vasopressor strategies for sepsis from intensive care records and reported that its selected treatments had higher estimated value than clinicians' decisions \cite{komorowski2018aiclinician}, and related studies applied reinforcement learning to sepsis management and mechanical ventilation \cite{raghu2017sepsis,peine2021ventilation}. Subsequent analyses and guidelines cautioned that off-policy value estimates from observational health data are sensitive to state definition, reward design, behaviour-policy estimation, and effective sample size, and questioned whether reported gains reflected genuinely better treatment \cite{gottesman2018evaluating,gottesman2019guidelines,jeter2019aiclinician,tang2021modelselection}. An early ``sepsis world model'' built from MIMIC data made these issues concrete: a learned variational autoencoder and mixture-density recurrent environment could be used to train a treatment policy, but because ground-truth counterfactuals were unavailable, the quality of that policy could not be verified within the simulator itself \cite{kiani2019sepsiswm}. More recent systems couple language-model agents with learned clinical world models in a propose--simulate--refine loop and report improved off-policy value and fewer unsafe actions on MIMIC-IV sepsis trajectories \cite{wu2026sepsisagent}; these remain retrospective estimates.

Clinical planning must also respect hard constraints. Constrained policy optimization and safety shields restrict actions to those that satisfy specified constraints \cite{achiam2017cpo,alshiekh2018shield}; in medicine, such constraints include dose ceilings, contraindications, physical force limits, and mandatory human handover. A general constrained planning objective can be written as
\begin{equation}
\bar a^{*}_{t:t+H-1}=\arg\max_{\bar a\in\mathcal{A}_{\mathrm{safe}}}\;\mathbb{E}_{p_\theta}\!\left[\sum_{k=t}^{t+H-1}\gamma^{\,k-t}\,r(s_k,a_k)\right]
\quad\text{s.t.}\quad
\Pr_{p_\theta}\!\left(c(s_k)>c_{\max}\right)\le\delta,\;\; k=t+1,\ldots,t+H,
\label{eq:constrained_planning}
\end{equation}
where \(p_\theta\) denotes the learned counterpart of the transition distribution in Eq.~(\ref{eq:dynamics}), \(r\) encodes clinical objectives, \(\gamma\) is a discount factor, \(c\) is a safety-relevant quantity such as toxicity or tissue force, \(\delta\) is a tolerated violation probability, and \(\mathcal{A}_{\mathrm{safe}}\) is the set of guideline- and device-admissible actions. Because the expectation and probability are evaluated under the learned model \(p_\theta\), any guarantee is only as reliable as the model's calibration in the region explored by the planner. Uncertainty-triggered abstention and clinician review are therefore integral components of planning-oriented medical world models rather than optional additions.

\subsection{A unified probabilistic formulation}\label{subsec:unified_formulation}

The components described above can be integrated within a partially observable Markov decision process \cite{kaelbling1998pomdp}. Let \(s_t\) denote the latent patient, procedural, or population state; \(a_t\) the clinical or device action; \(o_t\) the observations; and \(m_t\) an indicator of which observations were acquired. A medical world model then specifies
\begin{equation}
o_t\sim p_\theta(o_t\mid s_t,m_t),\qquad
m_t\sim p_\theta(m_t\mid s_t,a_{t-1}),\qquad
s_{t+1}\sim p_\theta(s_{t+1}\mid s_t,a_t),
\label{eq:pomdp}
\end{equation}
together with an inference model \(q_\psi(s_t\mid o_{\le t},m_{\le t},a_{<t})\) that maintains a belief over the current state, of which the encoder in Eq.~(\ref{eq:encoder}) is a simplified special case. Equation~(\ref{eq:pomdp}) makes three modelling commitments explicit: observations are noisy, partial views of the state; the decision to measure is itself informative; and transitions depend on actions. An \(H\)-step rollout samples
\begin{equation}
p_\theta(s_{t+1:t+H}\mid s_t,a_{t:t+H-1})=\prod_{k=t}^{t+H-1}p_\theta(s_{k+1}\mid s_k,a_k),
\label{eq:rollout}
\end{equation}
which can be decoded into future observations or clinical endpoints. The L1--L4 capability levels used in this Review correspond to progressively stronger uses of this structure. L1 applies Eq.~(\ref{eq:rollout}) with actions marginalized or fixed to observed care; L2 conditions transitions on specified actions; L3 compares rollouts under alternative action sequences, which requires the identification conditions associated with Eq.~(\ref{eq:gformula}); and L4 optimizes action sequences, as in Eq.~(\ref{eq:constrained_planning}). Each step upward adds assumptions that cannot be verified from predictive accuracy under observed care alone.

\subsection{Architectural evolution of world models}\label{subsec:architecture_evolution}

World models have evolved from environmental simulators towards medical dynamic modelling. In 2018, Ha and Schmidhuber proposed the World Models framework, which used a visual encoder, an MDN-RNN, and a controller to construct an internally rollable virtual environment \cite{ref082}. Dreamer then adopted and extended RSSM-based latent dynamics for imagined behaviour learning, substantially improving the efficiency of long-term simulation and planning \cite{ref002,ref005,ref016,ref008}. JEPA proposed predictive representation learning, reducing dependence on pixel-level reconstruction and strengthening high-level semantic modelling \cite{ref017,ref025,ref027}. Diffusion models further advanced high-fidelity dynamic simulation and showed potential for medical image generation and disease-course prediction \cite{ref039,ref014,ref020}. The development of multimodal foundation models and embodied intelligence has pushed world models to integrate vision, text, physiological signals, and clinical knowledge, forming medical world models that can support candidate long-term planning under clinical validation \cite{ref026,ref031,ref056,ref095,ref094,ref034,ref053,ref087,ref128}. Table~\ref{tab:technical_routes} compares these major technical routes and their medical-use boundaries.

Since 2023, several developments have broadened this design space further. Transformer-based world models such as IRIS represent environment dynamics as sequences of discrete tokens \cite{micheli2023iris}; diffusion world models such as DIAMOND indicate that preserving visual detail can matter for downstream control \cite{alonso2024diamond}; TD-MPC2 scales decoder-free latent dynamics with model-predictive control across diverse continuous-control tasks \cite{hansen2024tdmpc2}; V-JEPA 2 combines large-scale self-supervised video pretraining with an action-conditioned latent predictor for zero-shot robotic planning \cite{assran2025vjepa2}; and world foundation-model platforms such as Cosmos provide pretrained video world models that can be post-trained for specific physical domains \cite{nvidia2025cosmos}. Interactive generative environments, including Genie and UniSim, learn controllable simulators directly from video \cite{ref024,ref022}. Table~\ref{tab:general_wm} summarizes representative general-purpose architectures, their state and dynamics formulations, and the lessons and caveats they carry for medicine.

\begin{table}[!t]
\scriptsize
\setlength{\tabcolsep}{2pt}
\renewcommand{\arraystretch}{1.08}
\caption{Representative general-purpose world models and their implications for medicine. Years indicate first public release. The final two columns describe design lessons that have informed medical world models and the caveats that apply when these lessons are transferred to clinical settings.}\label{tab:general_wm}
\begin{tabularx}{\textwidth}{@{}>{\raggedright\arraybackslash}p{0.15\textwidth}>{\raggedright\arraybackslash}p{0.14\textwidth}>{\raggedright\arraybackslash}p{0.15\textwidth}>{\raggedright\arraybackslash}p{0.15\textwidth}YY@{}}
\toprule
Model (year) & State & Dynamics or generator & Action use & Lesson for medicine & Caveat\\
\midrule
Dyna (1991) \cite{sutton1991dyna} & Environment model & Learned transition model & Planning updates from simulated experience & Real and simulated experience can be combined & Assumes an accurate, low-dimensional model\\
World Models (2018) \cite{ref082} & Variational latent & Mixture-density recurrent network & Controller trained inside the learned model & Compact latent states can suffice for control & Policies can exploit model errors\\
PlaNet and Dreamer family (2019--2025) \cite{ref002,ref005,ref008,ref016} & Recurrent state-space latent & Deterministic and stochastic recurrent dynamics & Latent planning and actor--critic learning in imagination & Belief-state filtering suits partially observed patients & Compounding error over long horizons\\
MuZero (2020) \cite{ref004} & Value-equivalent latent & Learned dynamics predicting reward, value, and policy & Tree search & Planning may need only decision-relevant dynamics & Latent states are not clinically interpretable\\
IRIS (2022) \cite{micheli2023iris} & Discrete tokens & Autoregressive Transformer & Policy learning in imagination & Token sequences resemble tokenized health records & Tokenization may discard clinically salient detail\\
I-JEPA, V-JEPA, and V-JEPA 2 (2023--2025) \cite{ref017,ref027,assran2025vjepa2} & Predictive embedding & Latent prediction without pixel reconstruction; action-conditioned variant & Planning towards goal embeddings & Avoids modelling irrelevant appearance variation & Latent goals must be grounded in clinical targets\\
DINO-WM (2024) \cite{ref033} & Frozen pretrained features & Latent transition model & Model-predictive control in feature space & Pretrained medical encoders could define state spaces & Features may omit dynamics-relevant variables\\
TD-MPC2 (2023) \cite{hansen2024tdmpc2} & Decoder-free latent & Latent dynamics with value learning & Sampling-based model-predictive control & Scales across tasks; adopted for endovascular navigation & Reward design and transfer from simulation\\
DIAMOND (2024) \cite{alonso2024diamond} & Pixels & Diffusion model & Reinforcement learning inside the diffusion world & Fine visual detail, such as lesion margins, may be decision-relevant & Computational cost; hallucinated detail\\
GAIA-1, UniSim, and Genie (2023--2024) \cite{ref021,ref022,ref024} & Video tokens or latents & Generative video models & Text-, action-, or latent-action-conditioned interaction & Controllable simulators can be learned from video & Physical and causal fidelity are not guaranteed\\
Cosmos (2025) \cite{nvidia2025cosmos} & Tokenized video latent & Diffusion and autoregressive world foundation models & Post-training for domain-specific action conditioning & Basis for surgical world models and simulators & Domain shift from natural to clinical video\\
\botrule
\end{tabularx}
\end{table}

\subsection{Representative medical world models}
\label{subsec:representative_models}

Recent studies have applied world model formulations to longitudinal EHR modelling \cite{ref047,ref054}, medical imaging and future-observation modelling \cite{ref046,ref040,ref050}, treatment-response simulation \cite{ref039,ref070}, ultrasound guidance \cite{ref074,ref084}, and surgical interaction \cite{ref051}. Although these systems share an emphasis on modelling the evolution of clinical or procedural states, they differ in their state definitions, dynamics architectures, action or intervention semantics, prediction horizons, and validation settings \cite{ref056,ref095,ref094,ref034}. Their functional scope ranges from temporal or latent state prediction to action-conditioned simulation and comparison of candidate interventions. Table~\ref{tab:representative_models} compares selected models according to data modality, dynamics architecture, action or intervention representation, L1--L4 capability level, reported validation, and current evidence boundary.

\begin{table}[!t]
\scriptsize
\setlength{\tabcolsep}{1.0pt}
\renewcommand{\arraystretch}{1.0}
\caption{Representative medical world models, functional scope, and evidence boundaries. Comparison dimensions include data modality, dynamics architecture, action or intervention representation, L1--L4 capability level, and reported validation. The evidence boundary identifies the strongest available validation and the principal limitation on clinical interpretation. Examples span longitudinal electronic health records, medical imaging, treatment-response modelling, ultrasound guidance, and surgical interaction and are illustrative rather than exhaustive.}\label{tab:representative_models}
\begin{tabularx}{\textwidth}{@{}>{\raggedright\arraybackslash}p{0.15\textwidth}YYYYY@{}}
\toprule
Model & Data modality & Dynamics architecture & Action/intervention & Capability level & Validation and evidence boundary\\
\midrule
Cardiac Copilot [C] \cite{ref084} & Echo images and probe poses & Cardiac Dreamer latent spatial world model & Probe motion/pose & L2 & Standard-plane navigation on retrospective scans; no clinical-outcome evaluation\\
EHRWorld [P] \cite{ref047} & Longitudinal EHR events, notes, labs/vitals & LLM-based causal sequential conditional generator & Inquiries, medications, and procedures/interventions & L2 & Long-horizon in-hospital trajectory simulation; no explicit alternative-treatment comparison or established clinical-benefit evidence\\
EchoJEPA [P] \cite{ref046} & Echocardiography videos & JEPA-based latent predictive encoder & None & L1 & Representation transfer, EF estimation, view classification, and robustness\\
Brain-WM [P] \cite{ref050} & Longitudinal multi-sequence brain MRI and treatment history & Y-shaped MoT with treatment prediction and flow-based MRI generation & Treatment sequence/history & L2 & Future MRI and treatment prediction, including external cohorts; no explicit alternative-treatment comparison\\
CLARITY [P] \cite{ref070} & Longitudinal imaging with temporal and clinical context & Structured latent trajectory plus prediction-to-decision module & Treatment regimen/context & L3 & Candidate treatment planning in retrospective data; no prospective clinical-benefit evidence\\
Xray2Xray [P] \cite{ref040} & Multi-projection chest radiographs & Vision encoder plus latent projection-transition model & Projection angle/acquisition geometry & L2 (non-therapeutic) & Volumetric-context representation, synthesis, diagnosis, and risk prediction\\
EchoWorld [C] \cite{ref074} & Echo image--motion sequences & Motion-aware predictive pretraining and attention & Probe motion & L2 & Probe-guidance evaluation; embodied acquisition action rather than treatment\\
MeWM [C] \cite{ref039} & Pre/post-treatment CT and treatment protocols & VLM policy, tumour generative dynamics, and inverse-dynamics/survival evaluation & TACE treatment protocol & L3 & Simulated tumour evolution and candidate-protocol comparison; human second-reader setting, not clinical outcomes\\
SAW [P] \cite{ref051} & Laparoscopic video and tool trajectories & Trajectory-conditioned conditional video diffusion & Tool-action language, tool-tip trajectory, and tissue condition & L2 & Action-conditioned video generation and downstream data augmentation; no surgical-planning outcome validation\\
ChronoMedicalWorld [P] \cite{ref054} & Longitudinal EHR and structured/conversational interventions & Joint-embedding state encoder, wide action encoder, and recurrent latent transition & Structured interventions and patient--care communication & L2 & Long-horizon CKD/eGFR rollout; no explicit alternative-intervention comparison\\
\botrule
\end{tabularx}
\vspace{2pt}\par\scriptsize\textit{Note:} L1 indicates temporal or latent state prediction without explicit action conditioning;
L2, action-conditioned transition or generation;
L3, comparison of predicted outcomes under alternative actions;
and L4, closed-loop planning or control based on internal rollouts.
Capability levels describe functional scope rather than clinical evidence maturity.
\textit{Publication status:}
[C], peer-reviewed conference paper; [P], preprint.
\textit{Abbreviations:}
EHR, electronic health record;
LLM, large language model;
EF, ejection fraction;
MoT, Mixture of Transformers;
MRI, magnetic resonance imaging;
CT, computed tomography;
VLM, vision-language model;
TACE, transarterial chemoembolization;
CKD, chronic kidney disease;
eGFR, estimated glomerular filtration rate.
\end{table}
\FloatBarrier

\section{Applications, Evidence, and Clinical Translation}
\label{sec:applications}

Medical world models lie at the intersection of model-based reinforcement
learning, predictive representation learning, generative simulation,
medical foundation models, and causal treatment-effect estimation.
Dreamer-style latent imagination and JEPA-style predictive learning show
that a world model need not reconstruct every observable detail; prediction
and planning may instead operate on task-relevant latent states
\cite{ref001,ref082,ref005}. MuZero similarly demonstrated that
planning-relevant dynamics can be learned without explicit reconstruction
of the full environment, while DreamerV2 and DreamerV3 extended latent
rollout to increasingly complex visual-control settings
\cite{ref004,ref008,ref016}. More recent systems, including Genie, GAIA-1,
and interactive video simulators, have broadened this line of research
towards generative and interactive environments
\cite{ref028,ref021,ref022,ref024}. DINO-WM and related studies further
suggest that predicting future latent representations may be more useful
for planning and transfer than pixel-level reconstruction alone
\cite{ref025,ref027,ref033}. In medicine, these developments intersect
with causal methods for estimating longitudinal and individualized
treatment effects from observational patient data \cite{ref065}.

Medical world models are also closely related to foundation models, even
when the latter are not explicitly described as world models. Taming
Transformers and CLIP established scalable approaches to high-resolution
visual generation and language-supervised visual representation learning
\cite{ref009,ref010}. Related principles have since been adapted to
medical segmentation, retinal imaging, clinical language modelling, and
structured EHR representation learning
\cite{ref101,ref102,ref103,ref111}. Medical world models may therefore
build on pretrained imaging, EHR, language, and multimodal encoders rather
than learn patient-state representations entirely from scratch. Such
integration may improve data efficiency and multimodal coverage, but it
also introduces compound failure modes: representation error,
dynamics-model drift, planning bias, and human--AI interaction errors may
propagate across successive stages and become difficult to audit.

The emerging application landscape extends from static disease recognition
to future-state prediction, intervention-conditioned simulation, and
planning-oriented decision support. Unlike generative models concerned
primarily with observation synthesis, medical world models place greater
emphasis on how states evolve over time and under specified actions.
Current work can be organized into six broad, overlapping application
domains: medical imaging representation and future-observation simulation;
EHR and longitudinal patient-trajectory modelling; disease-progression and
treatment-response simulation; physiological-signal and multimodal-state
modelling; ultrasound guidance and surgical embodied interaction; and
population- or health-system simulation. These domains share a common focus on how current states evolve under
specific actions, but differ in state granularity, action semantics,
prediction horizon, and validation requirements.

\begin{figure}[!t]
\centering
\includegraphics[width=0.98\textwidth]
{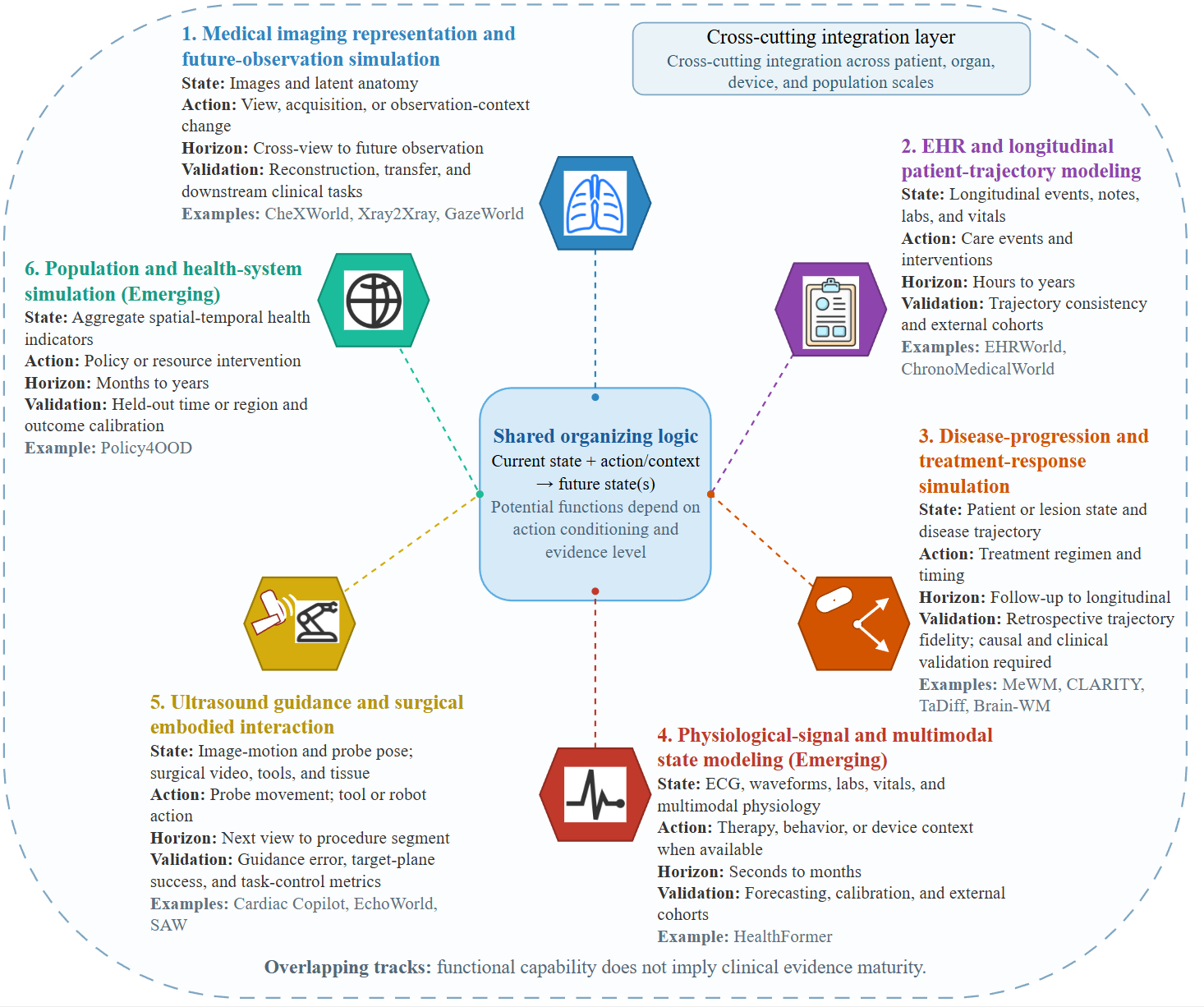}
\caption{Representative application domains of medical world models. Six overlapping domains are considered: medical imaging representation and simulation of future observations; longitudinal EHR and patient trajectory modelling; disease progression and treatment response simulation; physiological signal and multimodal state modelling; ultrasound guidance and surgical embodied interaction; and population and health system simulation. These domains share a common structure in which current states evolve under specified actions or contextual conditions, but differ in state definition, action semantics, prediction horizon, and validation requirements. The central integration layer connects information across patient, organ, device, institutional, and population scales and does not constitute an additional application category. Examples are illustrative rather than exhaustive.}
\label{fig:application_tracks}
\end{figure}

Figure~\ref{fig:application_tracks} summarizes the six domains and their
shared organizing framework. The central integration layer represents
connections across patient scales, devices, and population contexts rather
than an additional application category. Clinical digital twins are
similarly treated as a cross-cutting integration framework that may span
several domains, rather than as a separate seventh category. Recent
extensions include hierarchical action-guided modelling for fine-grained
auricular CT segmentation and belief-driven world modelling for sequential
2D--3D ultrasound registration
\cite{yang2026auricularworld,kang2026dreamreg}.

\subsection{Capability Levels and Claim-Specific Evidence}\label{subsec:capability_levels}

The L1--L4 framework proposed by Qazi and colleagues provides a useful
description of functional capability \cite{ref043}. L1 comprises temporal
prediction without an explicit action variable. L2 conditions state
transitions on specified actions. L3 compares predicted outcomes under
alternative actions that were not all observed for the same patient. L4
uses internal rollouts for closed-loop planning or control. To avoid
overstating capability, studies reporting action-conditioned generation
without explicit alternative-action comparison are classified as L2.
These levels describe functional scope rather than evidence quality or
clinical maturity. At the final search date of 20 July 2026, more than half
of the studies in the strict empirical subset were available as preprints,
including arXiv-only records. The following synthesis therefore maps the
technical landscape and its evidence gaps without treating methodological
novelty or recency as evidence of clinical reliability.

Each capability level corresponds to a different clinical claim and therefore to a different minimum evidence package (Table~\ref{tab:claims_evidence}). An L1 forecasting claim can be supported by temporally separated external validation and calibrated uncertainty. An L2 claim additionally requires evidence that predicted transitions change in the expected direction and magnitude when the action changes, which predictive accuracy under observed actions does not provide \cite{zhang2026mechanism}. An L3 claim requires an explicit causal estimand, identification assumptions, and positivity diagnostics, and an L4 claim requires safety-constrained evaluation, off-policy or prospective validation, and human oversight. Treating these levels as distinct evidential thresholds helps to avoid the common inference that a visually realistic or accurate rollout is sufficient to support treatment comparison.

\begin{table}[!t]
\scriptsize
\setlength{\tabcolsep}{2pt}
\renewcommand{\arraystretch}{1.08}
\caption{Capability levels, the clinical claims they support, and the minimum evidence required before clinical interpretation. Representative studies illustrate each level and are not exhaustive. Requirements are cumulative: a higher-level claim also requires the evidence listed for lower levels. Meeting the listed requirements is necessary but not sufficient for clinical use.}\label{tab:claims_evidence}
\begin{tabularx}{\textwidth}{@{}>{\raggedright\arraybackslash}p{0.05\textwidth}>{\raggedright\arraybackslash}p{0.16\textwidth}>{\raggedright\arraybackslash}p{0.13\textwidth}>{\raggedright\arraybackslash}p{0.17\textwidth}YY@{}}
\toprule
Level & Typical claim & Formal object & Representative studies & Appropriate metrics & Minimum evidence before clinical interpretation\\
\midrule
L1 & The state or its observations will evolve as forecast under usual care & Eq.~(\ref{eq:rollout}) with actions marginalized or fixed & EchoJEPA, CheXWorld, EyeWorld, SMB-Structure \cite{ref046,ref126,gao2026eyeworld,adam2026smbstructure} & Horizon-specific discrimination and calibration; trajectory error; image fidelity assessed with clinical reads & Temporal and external validation; calibration by horizon and subgroup; patient-level leakage controls \cite{ref066,vancalster2019calibration,subramanian2026leakage}\\
L2 & Under a specified action, the state will evolve as forecast & \(p_\theta(s_{t+1}\mid s_t,a_t)\) & Xray2Xray, EchoWorld, Brain-WM, SAW, ChronoMedicalWorld \cite{ref040,ref074,ref050,ref051,ref054} & Action-following fidelity, controllability, mechanism consistency & Tests of action faithfulness and known effect directions; documented action ontology \cite{zhang2026mechanism,wang2026bundles}\\
L3 & One action would lead to a better outcome than another for this patient & Contrast of rollouts approximating Eq.~(\ref{eq:gformula}) & CLARITY, MeWM, CaMeWM, ECG-WM \cite{ref070,ref039,gao2026camewm,chen2026ecgwm} & Estimated effect contrasts with uncertainty; policy value; agreement with trial evidence; sensitivity bounds & Prespecified estimand; target-trial emulation; overlap diagnostics; sensitivity analyses; benchmarking against randomized evidence \cite{hernan2016target,vanderweele2017evalue,qi2026causalaudit}\\
L4 & Following the model's plan improves outcomes safely & Eq.~(\ref{eq:constrained_planning}) & SepsisAgent; robotic ultrasound and endovascular planners \cite{wu2026sepsisagent,schueler2026usguidance,robertshaw2026safeendo} & Task success, constraint violations, override and abstention rates, off-policy value with confidence intervals & Safety case; staged prospective evaluation; post-deployment monitoring \cite{ref114,ref105,ref104,gottesman2019guidelines}\\
\botrule
\end{tabularx}
\end{table}

\subsection{Medical Imaging Representation and Future-Observation Simulation}\label{subsec:app_imaging}

Medical imaging provides a relatively well-defined setting for world-model
research because anatomical structure, acquisition viewpoint, temporal
follow-up, and device-domain variation offer identifiable states and
transition conditions. CheXWorld, Xray2Xray, and GazeWorld illustrate three
distinct directions: structured chest-radiograph representation,
projection-conditioned transformation, and modelling of radiologist
viewing behaviour \cite{ref040,ref055,ref126}. Their contribution lies
not simply in generating medical images, but in representing changes in
image domain, acquisition geometry, and information-acquisition behaviour
as learnable transition processes. Medical segmentation and retinal
foundation models further indicate that large-scale pretraining may reduce
the cost of downstream segmentation, detection, and cross-task transfer
\cite{ref101,ref103}.

CheXWorld does not generate post-treatment images. Instead, it uses
predictive learning to model local anatomical features, global spatial
organization, and domain variations associated with imaging devices and
institutions, and evaluates transfer to classification and segmentation
tasks \cite{ref126}. The model therefore applies world-model principles to
structured imaging representation, but its state space remains primarily
observational and contains no explicit therapeutic action. Under the
L1--L4 framework, it is consequently closer to L1.

Xray2Xray treats changes in radiographic projection as transitions within
the acquisition environment. Learning relationships across projection
views provides information about latent three-dimensional thoracic
structure and supports downstream disease classification,
cardiovascular-risk assessment, and volumetric reconstruction
\cite{ref040}. The acquisition angle can be represented as an action
condition, but it is an imaging action rather than a therapeutic
intervention. Accurate view transformation therefore does not imply the
ability to predict disease evolution under treatment.

GazeWorld extends the modelled state beyond the image to the
information-acquisition behaviour of radiologists. It represents the image
as an environment and the gaze sequence as a trajectory, predicting the
latent representation of the next attended region while estimating
information from unvisited regions \cite{ref055}. This formulation may
support diagnostic representation learning and reading-path assistance,
but gaze trajectories remain imperfect proxies for clinical reasoning.
Reproducing common viewing patterns does not establish diagnostic
correctness or generalizability across diseases, readers, and clinical
settings.

More recent studies extend these directions. X-WIN distils volumetric knowledge from CT into a chest-radiograph world model by learning to predict radiographic projections under 3D transformations in latent space \cite{yang2025xwin}, and DALE-CT reports that a depth-aware slice encoder trained with a joint-embedding objective develops an implicit anatomical trajectory through each CT volume without 3D or positional supervision \cite{damron2026dalect}. EyeWorld treats the eye as a partially observed dynamical system, learning a modality-stable ocular state with time-conditioned transitions for progression forecasting \cite{gao2026eyeworld}, and an MRI contrast-enhancement kinetics world model simulates continuous enhancement dynamics from sparsely sampled acquisitions \cite{kong2026cekworld}. Other systems use the language of world models for latent patient states that support both readout and conditional generation without modelling temporal transitions. MedDream learns a shared latent state from 2.65 million chest radiograph--text pairs and, in a reader study, increased mean resident concordance with radiologist consensus from 56.3\% to 63.0\% \cite{xi2026meddream}, whereas HounsWorld treats CT volumes and clinical language as observations of a shared state for readout, report reconstruction, and condition-specific CT simulation \cite{bai2026hounsworld}. EndoWorld-JEPA applies motion-guided latent prediction to endoscopic video pretraining \cite{wang2026endoworldjepa}. These studies broaden the range of imaging states that can be modelled, but most operate at L1 or at a non-therapeutic L2, and several do not model temporal transitions at all.

Taken together, imaging-oriented world models have relatively clear task
definitions and can be evaluated using classification, segmentation,
reconstruction, or view-synthesis metrics. Their clinical evidence,
however, remains limited. Most address how medical observations should be
represented when viewpoint, region, time, or imaging domain changes, rather
than how patients respond to alternative treatments. They may provide
useful components for subsequent disease-dynamics modelling, but do not by
themselves support therapeutic counterfactual simulation.

Generative imaging models also introduce a specific risk. Distribution-matching objectives can add or remove pathological features while preserving global realism \cite{cohen2018hallucinate}, and diffusion models used for longitudinal or counterfactual image generation inherit this vulnerability \cite{kazerouni2023diffusion}. Image-based models can likewise learn shortcuts, such as acquisition or institutional markers, that predict labels or outcomes without representing disease \cite{geirhos2020shortcut,degrave2021shortcuts,ref092}. Evaluation of imaging world models should therefore include reader studies focused on clinically salient findings, lesion-level rather than image-level similarity, and stress tests that perturb non-pathological factors such as scanner, site, and acquisition protocol.

\subsection{EHR and Longitudinal Patient-Trajectory Modelling}\label{subsec:app_ehr}

Longitudinal EHRs are the most direct source of patient-level state trajectories, because they record diagnoses, measurements, medications, procedures, and outcomes over months or years. Generative timeline models have progressed from task-specific prediction to sampling entire future trajectories. Foresight models patient timelines as sequences of biomedical concepts extracted from clinical text \cite{ref085}; ETHOS tokenizes patient health timelines for zero-shot prediction and simulation of health trajectories \cite{renc2024ethos}; CoMET shows that generative medical event models improve with scale \cite{ref041}; and Delphi-2M models the competing natural history of more than 1,000 diseases, with external validation in a separate national population \cite{shmatko2025delphi}. Clinical language models such as NYUTron further show that unstructured notes can support many prediction tasks and prospective deployment within a health system \cite{jiang2023nyutron}. These models capture rich temporal dependencies, but they typically treat treatments as tokens in the same stream as outcomes, leaving intervention semantics implicit.

Recent EHR world models make the state--action distinction explicit. In addition to EHRWorld and ChronoMedicalWorld \cite{ref047,ref054}, SMB-Structure combines next-token prediction with a joint-embedding objective so that future patient states must be predicted in latent space before they are observed \cite{adam2026smbstructure}. Clin-JEPA co-trains a language-model encoder with a latent trajectory predictor on hourly MIMIC-IV records and reports less degradation under 48-hour autoregressive rollout than two-stage designs \cite{yang2026clinjepa}. An action-conditioned EHR world model separates patient states from clinical actions through a state- and action-conditioned transition operator and shows that perturbing action sequences changes the generated trajectories \cite{lee2026ehrwm}, and PatientWM conditions each next-state prediction on an ordered history of latent states and actions, with evaluation across MIMIC-III, MIMIC-IV, and three external clinical datasets \cite{kan2026patientwm}. Large language models have also been probed as implicit medical world models by posing time-indexed questions about a patient's future from clinical documentation \cite{willems2026futurequerying}.

Two data problems recur in this domain. First, EHR events must be represented with the time at which information became available, not only the time to which it refers. EHR2Trace showed that assigning later diagnoses to admission time substantially inflated measured performance, and that a model trained on such data lost accuracy when deployed on availability-filtered histories \cite{yang2026ehr2trace}. Common data models such as OMOP and shared resources such as MIMIC and EHRSHOT improve reproducibility, but they do not by themselves resolve these temporal ambiguities \cite{hripcsak2015ohdsi,johnson2023mimiciv,ref019}. Second, the granularity of an intervention changes model behaviour. An audit of 945,707 patient-hours in MIMIC-IV found that some treatment components, such as the parameters of a dialysis circuit, never occur apart, and that editing a single ventilator setting moved the predicted next state less than editing the complete configuration recorded for a comparable patient \cite{wang2026bundles}. Counterfactual EHR simulation therefore requires action ontologies that reflect how care is actually delivered.

\subsection{Disease-Progression and Treatment-Response Simulation}\label{subsec:app_treatment}

Treatment-response simulation is among the most clinically consequential
and methodologically demanding applications of medical world models. MeWM,
CLARITY, TaDiff, and Brain-WM combine treatment conditions, follow-up time,
and longitudinal imaging within predictive or generative frameworks
\cite{ref039,ref070,ref020,ref050}. The central challenge is not simply
to generate a plausible future observation, but to estimate how the
patient trajectory would differ under an alternative intervention.
Counterfactual recurrent networks, Causal Transformer, and conditional
generative models provide methodological foundations for longitudinal
treatment-effect estimation, time-varying confounding adjustment, and
individualized treatment comparison
\cite{ref006,ref012,ref075}.

Causal machine-learning reviews, medical-imaging causality studies, and
real-world-evidence frameworks further indicate that clinically meaningful
counterfactual analysis requires explicit definitions of the intervention,
outcome, follow-up period, confounders, and target population
\cite{ref013,ref098,ref099}. Where appropriate, the analysis should also
specify the target trial or causal estimand being approximated. Visual
realism, image similarity, and trajectory consistency cannot by themselves
establish a causal treatment effect. Accordingly, consistency metrics
reported by MeWM, or similar generative systems, should not be interpreted
as evidence that one treatment improves clinical outcome.

Several recent studies extend treatment-response modelling. CaMeWM formulates post-TACE treatment planning as a support-restricted g-formula problem: queried regimens are admitted only for clinically comparable histories with empirical treatment support, lesion-level follow-up CT is generated through an action-conditioned diffusion transition, and survival utility of the induced policy is evaluated with censoring-adjusted doubly robust estimation across 79 out-of-fold episodes \cite{gao2026camewm}. CoRe-WM conditions latent transitions of breast dynamic contrast-enhanced MRI on patient profile and assigned therapy, reads out pathological complete response at successive decision points, and reports cross-cohort transfer from I-SPY2 to I-SPY1 \cite{zhao2026corewm}. An intervention-aware cardiology world model updates a 3D latent cardiac state through time-ordered post-ablation events to forecast atrial fibrillation recurrence \cite{chung2026afablation}, and an immune world model learns how perturbations move immune states across cellular, tissue, and individual scales \cite{cui2026immunewm}. Equally important are audits of what such models actually learn. The Patient, Place, Prior (P\textsuperscript{3}) framework asks whether a forecast benefits from a patient's own longitudinal history beyond a population-average prediction; in its breast MRI example, the descriptive interval for patient-specific benefit included zero \cite{gu2026p3}.

Mechanistic and hybrid approaches remain an important counterpart. Image-calibrated reaction--diffusion models of glioma and related oncology digital twins use few parameters with direct biological meaning and can be calibrated to individual patients \cite{swanson2000glioma,hormuth2021glioma,wu2022oncologytwin}, and conditional generative models have simulated detailed individual trajectories of Alzheimer's disease progression \cite{fisher2019adforecast}, an approach proposed for generating digital twins of clinical-trial participants \cite{bordukova2024gendt}. Learned world models can borrow from these traditions by using mechanistic models as priors, as constraints on admissible transitions, or as independent checks on rollouts.

\subsection{Physiological-Signal and Multimodal-State Modelling}\label{subsec:app_physiology}

Physiological signals provide dense, high-frequency observations of state and, particularly in critical care, are frequently paired with well-documented interventions. Open intensive care databases such as MIMIC-III, MIMIC-IV, eICU, HiRID, and AmsterdamUMCdb provide time series of vital signs, laboratory values, and treatments \cite{johnson2016mimiciii,johnson2023mimiciv,pollard2018eicu,hyland2020hirid,thoral2021amsterdam}, and have supported early-warning models, some with external validation, as well as community benchmarks such as the PhysioNet sepsis challenge \cite{hyland2020hirid,reyna2020sepsis,goldberger2000physionet}. Within our evidence map, HealthFormer models multi-domain physiological trajectories from a deeply phenotyped multi-visit cohort and, in a held-out personalized-nutrition trial, used intervention-conditioned predictions to recover individual six-month biomarker changes \cite{ref052}.

World-model formulations are now being applied directly to physiological signals. ECG-WM integrates physiological ordinary-differential-equation priors into latent diffusion dynamics to simulate post-intervention ECG trajectories and uses sampling variability for uncertainty-aware comparison of candidate interventions \cite{chen2026ecgwm}. NormWear-2 encodes multivariate physiological signals and intervention variables in a shared latent space for forecasting across daily-life, point-of-care, and clinical settings \cite{luo2026normwear2}. PHYSIO WORLD constrains the direction of drug effects using 70 pharmacological rules covering 28 drugs and reported the lowest out-of-distribution error across five distribution-shift settings in cohorts from three intensive care databases \cite{xu2026physioworld}. Action- or event-conditioned joint-embedding models have also been proposed for cardiac electrophysiological dynamics and wearable inertial kinematics \cite{fernandes2026cardiacjepa,delaney2026sonata}, and a conceptual framework for physiological world models defines capability levels and benchmark tasks for whole-person state transitions \cite{zhang2026pwm}.

Physiological simulation has a longer history in diabetes. The UVA/Padova type 1 diabetes simulator, built from a cohort of virtual subjects spanning observed metabolic variability, was accepted by the US Food and Drug Administration as a substitute for animal trials in the preclinical testing of closed-loop control strategies \cite{kovatchev2009insilico,man2014uvapadova}. This precedent shows that simulators can earn a regulatory role, but only for a narrowly defined purpose and after extensive validation against clinical data. Learned physiological world models will need comparable, purpose-specific credibility evidence.

\subsection{Ultrasound Guidance and Surgical Embodied Interaction}\label{subsec:app_embodied}

Embodied settings are the closest medical analogue of the robotic environments in which world models were developed: the action is a probe, instrument, or catheter motion; the observation is an image stream; and the consequences of an action become visible within seconds. In echocardiography, Cardiac Copilot and EchoWorld learn probe-motion-conditioned dynamics for standard-plane navigation \cite{ref084,ref074}, and a structure-aware world model pretrained on more than 1.36 million echocardiograms with 3D spatial poses reduced probe-guidance errors across ten standard views \cite{jiang2024structurewm}. DreamReg extends belief-driven world modelling to sequential 2D--3D ultrasound registration \cite{kang2026dreamreg}. Newer systems move towards closed-loop planning. A retrieval-induced latent transition model built on V-JEPA 2.1 features reached the target carotid view in 86\% of retrospective closed-loop episodes and in 83\% of attempts in a prospective feasibility study on unseen volunteers \cite{schueler2026usguidance}; a scene-graph world model planned robotic probe trajectories and reached the target view in 73.7\% of robot--phantom trials \cite{li2026sonograph}; and an action-conditioned diffusion world model provided rewards for fine-tuning a goal-conditioned policy for robotic carotid and thyroid scanning \cite{fan2026neckus}. SonoWorld and UltraWorld instead address the scarcity of tracked data by learning pose-controllable ultrasound generators primarily from untracked clinical videos \cite{yang2026sonoworld,yang2026ultraworld}. These advances build on earlier deep reinforcement learning for probe navigation and on the broader development of robotic ultrasound \cite{li2021usnavigation,jiang2023roboticus}.

Surgical world models have progressed from action-conditioned video generation towards policy learning. Early work used world-model-based deep reinforcement learning for visuomotor surgical grasping in simulation and on real robots, including phantom and ex vivo settings \cite{lin2024gas,lin2025gasv2}, while video diffusion models were fine-tuned to generate fine-grained suturing sub-actions \cite{turkcan2025suturing} and video generation models were explored as endoscopy simulators \cite{li2024endora}. SAW and the Surgical Vision World Model condition video generation on tool actions or inferred latent actions \cite{ref051,ref036}; SurgVista targets long-horizon instrument--tissue consistency \cite{pan2026surgvista}; SurgCast uses predicted instrument skeletons as a control interface \cite{liu2026surgcast}; Surg-UniWorld composes multimodal visual control signals \cite{zhou2026surguniworld}; CrossScope jointly predicts video from two cooperating scopes in endoscopic retrograde cholangiopancreatography (ERCP) \cite{liu2026crossscope}; and SWoMo decouples symbolic motion simulation from diffusion-based appearance in cataract surgery \cite{sivakumar2026swomo}. At larger scale, the Open-H-Embodiment consortium assembled video with synchronized kinematics from more than 50 institutions and multiple robotic platforms, and trained both a vision--language--action policy and an action-conditioned, multi-embodiment surgical world model \cite{openh2026embodiment}. Cosmos-H-Surgical generates surgical videos from which pseudo-kinematics are inferred to augment policy training \cite{he2025cosmoshsurgical}, and Cosmos-H-Dreams distils such a model into a real-time simulator that can be driven by human operators, a commercial surgical console, or learned policies \cite{tejero2026cosmoshdreams}. Surgical WAM jointly predicts future endoscopic observations and robot action chunks and reported that action-free video pretraining improved success in simulated tasks \cite{bao2026surgicalwam}; a hierarchical event-structured world model addresses long-horizon manipulation in simulation \cite{zhang2026s2hwm}; and an action-conditioned flow-matching world model has been used to detect failures of surgical imitation policies on the da Vinci Research Kit, with alarm thresholds set by conformal calibration \cite{huang2026fomofd,kazanzides2014dvrk}.

Endovascular navigation has emerged as a further testbed. TD-MPC2-based world models trained across patient-specific vasculatures outperformed a model-free baseline in simulation and were evaluated in vitro under fluoroscopy \cite{robertshaw2025mtwm,robertshaw2026safeendo,robertshaw2026pef}, whereas a belief-state formulation with a physics-based catheter model explored navigation under deliberately sparse fluoroscopy in a planar simulated proof of concept \cite{rijhwani2026catheter}. EndoWAM applies a world--action model to robotic endoscopic navigation across ureteroscopy, oesophagoscopy, and ERCP \cite{lin2026endowam}.

Evaluation is the principal bottleneck in this domain. Generation metrics such as Fr\'echet video distance do not measure whether predicted instrument motion is geometrically accurate or actionable, which motivated motion-centric benchmarks such as SurgWMBench \cite{liu2026surgwmbench}. When four board-certified surgeons rated zero-shot surgical video generated by a state-of-the-art video model, they found high visual plausibility but critical failures in instrument operation, tissue feedback, and surgical intent \cite{ref083}; similarly, the REMEDY benchmark found that even the best evaluated generator achieved only 28.25\% strict clinical success on medical teaching demonstrations \cite{tan2026remedy}. Embodied evidence for medical world models remains largely confined to simulation, phantoms, ex vivo tissue, and volunteers. By comparison, autonomous robotic anastomosis has been demonstrated in vivo in porcine models using planning and tracking without a learned world model \cite{saeidi2022anastomosis}, and imitation-learning policies such as the Surgical Robot Transformer have been trained on da Vinci Research Kit demonstrations \cite{kim2024srt}. Shared platforms and the broader surgical data science agenda provide infrastructure for progressing from simulated to clinical evaluation \cite{xu2021surrol,kazanzides2014dvrk,maierhein2017sds}.

\subsection{Population and Health-System Simulation}\label{subsec:app_population}

At population and system scales, the state may comprise disease prevalence, care capacity, or the joint distribution of many individual trajectories, and actions may be policies, resource allocations, or outreach campaigns. Within our evidence map, Policy4OOD and world-model-enhanced offline reinforcement learning address policy-intervention simulation and sequential intervention optimization \cite{ref048,ref049}. Recent perspectives argue that population health and epidemiology are natural settings for world models, because the true burden of disease is latent, surveillance signals are delayed and policy-dependent, and interventions act through behavioural and social feedback \cite{xie2026phwm,memon2026epiwm}. Established agent-based epidemic models such as Covasim, and discrete-event simulations of hospital operations, provide mechanistic baselines against which learned models should be compared \cite{kerr2021covasim,gunal2010des}, and synthetic-population generators such as Synthea supply controlled test beds \cite{walonoski2018synthea}.

Language-model agents add another form of system simulation. Generative agents can produce believable simulations of human behaviour in interactive environments \cite{park2023generative}; Agent Hospital simulates hospital workflows with evolvable medical agents \cite{li2024agenthospital}; and AgentClinic and MedAgentBench evaluate clinical agents in simulated consultations and virtual EHR environments \cite{schmidgall2024agentclinic,jiang2025medagentbench}. These environments are useful for testing workflows and agent behaviour, but their patients and populations are themselves model-generated and need not reflect real disease dynamics. A patient world model for direct-to-consumer health campaigns illustrates the inferential risk: switching off all future exposure increased predicted conversion from 0.31 to 0.89, a pattern the authors attributed to selection effects in observational exposure data rather than to a causal effect \cite{wang2026dtcwm}. System-level simulators therefore require the same separation between forecasting and intervention claims as patient-level models.

\subsection{Clinical Digital Twins as a Cross-Cutting Integration Framework}\label{subsec:app_digitaltwin}

Clinical digital twins are virtual representations of an individual patient, organ, procedure, or care system that are updated with data from their physical counterpart and used to inform decisions \cite{ref110,ref117,venkatesh2022healthdt}. They predate learned world models. Cardiac digital twins combine patient-specific anatomy and electrophysiology with biophysical simulation \cite{corral2020digitaltwin,gillette2021cardiactwin}, a roadmap has been proposed for digital twins of the human immune system \cite{laubenbacher2022immune}, and oncology twins calibrate tumour-growth models to serial imaging \cite{wu2022oncologytwin}. Scaling such twins from bespoke research models to routine use requires automated personalization, credibility assessment, and supporting infrastructure \cite{niederer2021scaling}, and the US National Academies identified verification, validation, and uncertainty quantification as foundational research needs \cite{nasem2024digitaltwins}.

Medical world models can serve as the learned predictive core of a digital twin, providing state estimation and intervention-conditioned rollout where mechanistic knowledge is incomplete, while mechanistic components provide structure where data are sparse \cite{zhou2026digitaltwinai,wang2026biomedwm}. Language-model-based approaches such as TWIN-GPT have been proposed to generate digital twins of clinical-trial participants \cite{wang2024twingpt}, and generative models are being explored more broadly for trial design and synthetic control arms \cite{bordukova2024gendt}. A digital twin also introduces obligations beyond those of a stand-alone model, including synchronization with new data, version control, ethical questions regarding personal data and responsibility, and continuous revalidation as the physical counterpart changes \cite{bruynseels2018ethics}. For these reasons, and consistent with Figure~\ref{fig:application_tracks}, we treat digital twins as an integration framework spanning the six application domains rather than as a separate category.

\subsection{Recent Studies and Emerging Directions}\label{subsec:horizon}

Tables~\ref{tab:horizon_patient} and~\ref{tab:horizon_embodied} summarize representative recent studies. Three patterns emerge. First, activity has increased sharply: most listed studies were released in 2026. Second, there is a visible shift from L1 representation learning towards L2 action conditioning and, in embodied domains, towards L4 closed-loop control in simulation, phantoms, or volunteers. Third, a parallel body of methodological work has begun to audit medical world models directly, examining patient-level leakage, personalization, mechanism consistency, counterfactual harm bounds, calibration under shift, and federated training \cite{subramanian2026leakage,gu2026p3,zhang2026mechanism,qi2026causalaudit,khan2026caltwin,wei2026federatedwm}. Because most of these studies are preprints with retrospective, simulated, or preclinical evaluations, the tables should be read as a map of research activity rather than as evidence of clinical performance.

\begin{table}[p]
\scriptsize
\setlength{\tabcolsep}{2pt}
\renewcommand{\arraystretch}{1.0}
\caption{Representative recent studies of patient-level medical world models in imaging, longitudinal EHR, treatment-response, and physiological domains. Capability levels follow the L1--L4 definitions in Table~\ref{tab:representative_models}, and the final column summarizes the principal reported evaluation of each study. These studies are not part of the strict empirical subset.}\label{tab:horizon_patient}
\begin{tabularx}{\textwidth}{@{}>{\raggedright\arraybackslash}p{0.13\textwidth}>{\raggedright\arraybackslash}p{0.15\textwidth}>{\raggedright\arraybackslash}p{0.17\textwidth}Y>{\raggedright\arraybackslash}p{0.07\textwidth}Y@{}}
\toprule
Study & Domain and data & State; action or condition & Mechanism & Level & Reported evaluation\\
\midrule
X-WIN [P] \cite{yang2025xwin} & Chest radiography with CT-derived projections & 3D-aware radiograph state; projection under 3D transformation & Latent projection prediction with contrastive alignment & L2\textsuperscript{*} & Linear-probe and few-shot transfer; CT reconstruction from projections\\
DALE-CT [P] \cite{damron2026dalect} & Chest CT & Slice-level anatomical trajectory; none & Joint-embedding objective with depth-aware slab sampling & L1 & Classification on CT-RATE; external transfer\\
EyeWorld [P] \cite{gao2026eyeworld} & Multimodal ophthalmic imaging & Modality-stable ocular state; elapsed time & Generative latent state with time-conditioned transitions & L1 & Parsing, cross-modality translation, progression forecasting\\
MRI CEKWorld [P] \cite{kong2026cekworld} & Contrast-enhanced MRI & Enhancement state; time after contrast & Latent alignment and difference learning & L1 & Content realism and kinetic continuity on two datasets\\
MedDream [P] \cite{xi2026meddream} & 2.65 million radiograph--text pairs & Shared radiographic state; report or subgroup condition & Joint understanding and generation & L1\textsuperscript{\dag} & Eight datasets; reader study of resident concordance\\
HounsWorld [P] \cite{bai2026hounsworld} & CT and clinical language & Shared patient state; imaging condition & Multimodal joint understanding and generation & L1\textsuperscript{\dag} & Readout, reconstruction, and simulation benchmark\\
SMB-Structure [P] \cite{adam2026smbstructure} & Structured EHR in oncology and pulmonary embolism cohorts & Latent patient state; none explicit & Joint-embedding prediction with next-token supervision & L1 & Linear probes along disease trajectories in two cohorts\\
Clin-JEPA [P] \cite{yang2026clinjepa,wang2026bundles} & Hourly MIMIC-IV records as text & Language-model latent state; hourly treatment text & Co-trained encoder and latent predictor & L2 & 48-hour rollout stability; 34 downstream tasks; treatment-bundle edits\\
Action-conditioned EHR world model [P] \cite{lee2026ehrwm} & Longitudinal EHR & Latent patient state; observed clinical actions & State-dependent linearized transition operator & L2 & Rollout fidelity; classification and time-to-event tasks; action sensitivity\\
PatientWM [P] \cite{kan2026patientwm} & MIMIC-III, MIMIC-IV, and three external datasets & Unified patient state with latent history; clinical actions & Availability-aware encoders and history-conditioned transitions & L2 & Prediction, multi-step forecasting, and action-planning evaluation\\
CaMeWM [P] \cite{gao2026camewm} & Pre- and post-TACE CT & Lesion-local state; TACE regimen & Support-restricted g-formula with 3D diffusion transition and survival utility & L3 & Follow-up CT similarity; survival concordance; doubly robust policy evaluation\\
CoRe-WM [P] \cite{zhao2026corewm} & Breast DCE-MRI (I-SPY2, I-SPY1) & Response state; therapy and patient profile & Conditioned latent transition with factorized response pathway & L2 & Pathological complete response discrimination; cross-cohort transfer\\
Post-ablation cardiology world model [P] \cite{chung2026afablation} & Cardiac MRI and post-procedure records & 3D latent cardiac state; post-procedure events & Event-driven latent updates & L2 & Recurrence prediction in internal cross-validation\\
ECG-WM [P] \cite{chen2026ecgwm} & ECG under pharmacological exposure & Electrophysiological state; drug intervention & ODE-regularized latent diffusion & L3 & Drug-response scenarios and clinical records; risk calibration\\
NormWear-2 [P] \cite{luo2026normwear2} & Physiological signals from 8,026 subjects & Shared physiological state; intervention variables & Pretrained dynamics with non-parametric adaptation & L2 & Multi-scale forecasting across four settings\\
PHYSIO WORLD [P] \cite{xu2026physioworld} & Three intensive care databases & Vital-sign state; drug doses & Sign-constrained intervention effects & L2 & Out-of-distribution error across five shift settings\\
SepsisAgent [P] \cite{wu2026sepsisagent} & MIMIC-IV sepsis trajectories & Clinical world-model state; fluids and vasopressors & Language-model agent with propose--simulate--refine loop and model-based reinforcement learning & L4 & Off-policy value; guideline adherence; unsafe actions\\
Immune world model [P] \cite{cui2026immunewm} & Cellular, tissue, and patient immune data & Multiscale immune state; perturbations & Action-conditioned model constructed by an automated scientific pipeline & L3 & Unseen interventions and combinations; therapeutic hypothesis generation\\
\botrule
\end{tabularx}
\vspace{2pt}\par\scriptsize\raggedright\textit{Note:} [P], preprint (publication status of the cited version). \textsuperscript{*}The action is a non-therapeutic acquisition or viewpoint change. \textsuperscript{\dag}Learns a latent state used for readout or conditional generation but does not model temporal transitions. Abbreviations: TACE, transarterial chemoembolization; DCE-MRI, dynamic contrast-enhanced MRI; ODE, ordinary differential equation.
\end{table}

\begin{table}[p]
\scriptsize
\setlength{\tabcolsep}{2pt}
\renewcommand{\arraystretch}{1.0}
\caption{Representative recent studies of embodied and system-level medical world models in ultrasound, surgery, endovascular and endoscopic navigation, and population settings. Capability levels follow the L1--L4 definitions in Table~\ref{tab:representative_models}, and the final column summarizes the principal reported evaluation of each study. These studies are not part of the strict empirical subset.}\label{tab:horizon_embodied}
\begin{tabularx}{\textwidth}{@{}>{\raggedright\arraybackslash}p{0.14\textwidth}>{\raggedright\arraybackslash}p{0.15\textwidth}>{\raggedright\arraybackslash}p{0.15\textwidth}Y>{\raggedright\arraybackslash}p{0.08\textwidth}Y@{}}
\toprule
Study & Domain and data & State; action & Mechanism & Level & Reported evaluation\\
\midrule
Structure-aware probe-guidance model [P] \cite{jiang2024structurewm} & Echocardiography with 3D poses & Cardiac structure; probe pose & Masked-structure prediction and pose-conditioned plane imagination & L2 & Reduced guidance error across ten standard views\\
SonoWorld [P] \cite{yang2026sonoworld} & Multi-organ ultrasound video; tracked thyroid scans & Ultrasound frames; 6-DoF probe pose & Latent-action video diffusion grounded to metric poses & L2 & Pose-following fidelity; user study; model-based action discrimination\\
UltraWorld [P] \cite{yang2026ultraworld} & Untracked clinical ultrasound video & Local observations; probe pose and imaging settings & Self-distilled action-conditioned generator & L4 (simulated) & Simulated closed-loop local planning versus visual servoing\\
Retrieval-based operator guidance [P] \cite{schueler2026usguidance} & Carotid ultrasound & Latent view state; probe position and orientation & Non-parametric retrieval transitions with receding-horizon planning & L4 & Retrospective closed loop; prospective feasibility in volunteers\\
SonoGraph-WM [P] \cite{li2026sonograph} & CT-derived scene graphs; robot and phantom & Anatomical scene graph; probe pose & Transformer predicting graphs and poses with receding-horizon planning & L4 & Simulated navigation; robot--phantom target reach\\
Neck ultrasound world model [P] \cite{fan2026neckus} & Robotic carotid and thyroid ultrasound & Ultrasound frames; probe motion & Latent diffusion world model providing rewards to a goal-conditioned policy & L4 & Real-world closed-loop guidance\\
GAS [C] and GASv2 [P] \cite{lin2024gas,lin2025gasv2} & Surgical grasping in simulation, phantom, and ex vivo settings & Depth and mask images; gripper motion & World-model-based deep reinforcement learning with sim-to-real transfer & L4 & Grasping success on real robots, including unseen objects\\
Cosmos-H-Surgical [P] \cite{he2025cosmoshsurgical} & Surgical video with action descriptions & Video; action text & Video world model with inverse dynamics for pseudo-kinematics & L2 & Policy trained with synthetic data evaluated on a surgical robot\\
Open-H-Embodiment [P] \cite{openh2026embodiment} & Video and kinematics from more than 50 institutions & Video; robot kinematics & Multi-embodiment action-conditioned simulator and policy & L2 & In silico policy evaluation; synthetic data generation\\
Cosmos-H-Dreams [P] \cite{tejero2026cosmoshdreams} & Open-H-Embodiment and procedure-specific data & Video; robot kinematics & Distilled causal few-step video generator & L2 & Real-time control by humans, a surgical console, and policies\\
SurgVista [P] \cite{pan2026surgvista} & Surgical video & Frames; instrument actions & Deformation-consistency and drift-adaptation training & L2 & Long-horizon interaction fidelity on a dedicated benchmark\\
SurgCast [P] \cite{liu2026surgcast} & Two robotic surgery video datasets & Instrument skeletons; robot actions & Skeleton forecasting to control video generation & L2 & Video quality and instrument consistency\\
Surgical WAM [P] \cite{bao2026surgicalwam} & Simulated surgical manipulation & Endoscopic frames; action chunks & Joint video and action prediction with receding-horizon execution & L2\textsuperscript{\S} & Simulated task success with and without video pretraining\\
FoMo-FD [P] \cite{huang2026fomofd} & da Vinci Research Kit, simulation and real & Latent frames; robot actions & Action-conditioned flow matching with conformal alarms & L2 & Failure detection versus false-alarm rate\\
SWoMo [P] \cite{sivakumar2026swomo} & Cataract surgery & Scene graph; surgical actions & Rule-based simulator with diffusion appearance & L2 & Generalization to unseen interaction geometries; downstream phase detection\\
Endovascular TD-MPC2 agents [C, P] \cite{robertshaw2025mtwm,robertshaw2026safeendo,robertshaw2026pef} & Patient-specific vasculatures; in vitro phantoms & Device and vessel state; catheter commands & Latent model-predictive control & L4 & Success versus a model-free baseline in silico; in vitro fluoroscopic evaluation\\
EndoWAM [P] \cite{lin2026endowam} & Ureteroscopy, oesophagoscopy, ERCP & Endoscopic frames; navigation actions & Future target grounding with an action expert & L2\textsuperscript{\S} & Generalization to unseen viewpoints, environments, and targets\\
Campaign patient world model [P] \cite{wang2026dtcwm} & 147,173 patients in a health campaign & Latent patient state; campaign exposure & Exposure-conditioned dynamics with a conversion hazard & L2 & Prescription-volume forecasting; scenario simulation flagged as confounded\\
\botrule
\end{tabularx}
\vspace{2pt}\par\scriptsize\raggedright\textit{Note:} [C], peer-reviewed conference paper; [P], preprint (publication status of the cited version). \textsuperscript{\S}World--action models that jointly predict observations and actions are classified as L2 unless internal rollouts are used to evaluate alternative actions. Abbreviations: 6-DoF, six degrees of freedom; ERCP, endoscopic retrograde cholangiopancreatography.
\end{table}

\subsection{Data Resources and Benchmarks}\label{subsec:data_resources}

Progress in medical world models depends on data that record states, actions, and outcomes over time at sufficient resolution. Table~\ref{tab:datasets} summarizes representative resources. Critical care databases provide the richest pairing of high-frequency states with documented treatments \cite{johnson2016mimiciii,johnson2023mimiciv,pollard2018eicu,hyland2020hirid,thoral2021amsterdam}; population cohorts and registries provide long horizons but sparse intervention detail \cite{sudlow2015ukbiobank,shmatko2025delphi}; longitudinal imaging cohorts such as ADNI and LUMIERE support disease-progression and treatment-response modelling \cite{petersen2010adni,suter2022lumiere}; echocardiography, chest-radiograph, and surgical-video datasets support representation and embodied modelling but rarely include patient outcomes \cite{ouyang2020echonet,leclerc2019camus,johnson2019mimiccxr,irvin2019chexpert,twinanda2017endonet,nwoye2022rendezvous}; and robotic datasets and simulators supply synchronized actions \cite{openh2026embodiment,xu2021surrol,kazanzides2014dvrk}. Few resources simultaneously provide explicit action semantics, long-horizon outcomes, and multi-institutional coverage, and none was designed to evaluate counterfactual claims. Benchmarks are beginning to target world-model-specific capabilities, including environment-level world-model queries, surgical motion planning, medical education video generation, and auditable EHR event construction \cite{ref064,liu2026surgwmbench,tan2026remedy,yang2026ehr2trace}. Evaluation of L3 and L4 claims, however, still relies on semi-synthetic data, off-policy estimates, or simulators whose own validity must be established \cite{man2014uvapadova,walonoski2018synthea}.

\begin{table}[!t]
\scriptsize
\setlength{\tabcolsep}{2pt}
\renewcommand{\arraystretch}{1.05}
\caption{Representative data resources and simulators for medical world models. The table indicates the state and action information each resource provides, its typical use in world-model research, and its principal limitation. Inclusion does not imply suitability for counterfactual or planning claims.}\label{tab:datasets}
\begin{tabularx}{\textwidth}{@{}>{\raggedright\arraybackslash}p{0.14\textwidth}>{\raggedright\arraybackslash}p{0.16\textwidth}YYYY@{}}
\toprule
Resource & Setting & State information & Action information & World-model use & Principal limitation\\
\midrule
MIMIC-III and MIMIC-IV \cite{johnson2016mimiciii,johnson2023mimiciv} & Single-centre intensive care and hospital EHR & Vital signs, laboratory values, notes, diagnoses & Medications, fluids, vasopressors, ventilation, procedures & ICU trajectory and treatment-response modelling; offline reinforcement learning & Single centre; documentation-driven timing; confounding by indication\\
eICU-CRD \cite{pollard2018eicu} & Multi-centre intensive care database & Vital signs, laboratory values, care plans & Treatments and interventions & External validation across hospitals & Heterogeneous documentation\\
HiRID and AmsterdamUMCdb \cite{hyland2020hirid,thoral2021amsterdam} & European intensive care databases & High-resolution physiology and laboratory values & Drug administration, ventilation & Fine-grained physiological dynamics; cross-site validation & Single-centre sources; access agreements\\
PhysioNet and the 2019 sepsis challenge \cite{goldberger2000physionet,reyna2020sepsis} & Physiological signal repository and multi-hospital challenge data & Signals, hourly vital signs and laboratory values & Limited & Forecasting and early-warning benchmarks & Few explicit actions\\
UK Biobank \cite{sudlow2015ukbiobank} & Population cohort of about 500,000 adults & Baseline measurements, imaging, linked health records & Sparse & Long-horizon natural-history modelling & Healthy-volunteer selection; irregular follow-up\\
ADNI \cite{petersen2010adni} & Multi-site Alzheimer's disease cohort & Longitudinal imaging, cognition, biomarkers & Largely observational & Disease-progression modelling & Selected research population\\
LUMIERE \cite{suter2022lumiere} & Longitudinal glioblastoma MRI with expert response assessment & Serial multi-sequence MRI & Treatment context & Treatment-response and growth modelling & Small cohort\\
MIMIC-CXR and CheXpert \cite{johnson2019mimiccxr,irvin2019chexpert} & Large chest-radiograph collections & Radiographs, reports, labels & None & Pretraining of imaging world models & Largely cross-sectional; label noise\\
EchoNet-Dynamic and CAMUS \cite{ouyang2020echonet,leclerc2019camus} & Echocardiography video datasets & Cardiac motion and function & None & Predictive representation learning & No probe kinematics or outcomes\\
Cholec80 and CholecT50 \cite{twinanda2017endonet,nwoye2022rendezvous} & Laparoscopic cholecystectomy video & Phases, tools, action triplets & Annotated actions without kinematics & Action-conditioned surgical video generation & Single procedure type; no robot kinematics\\
Open-H-Embodiment \cite{openh2026embodiment} & Medical robot video from more than 50 institutions & Surgical, ultrasound, and endoscopy video & Synchronized robot kinematics & Multi-embodiment world models and policies & Clinical outcomes not recorded\\
SurRoL and dVRK \cite{xu2021surrol,kazanzides2014dvrk} & Simulation platform and open research robot & Simulated or real robot states & Robot commands & Closed-loop policy learning and evaluation & Simulation-to-reality gap\\
Synthea and UVA/Padova simulator \cite{walonoski2018synthea,man2014uvapadova} & Synthetic patients; physiological simulator & Simulated records or glucose dynamics & Simulated treatments such as insulin & Controlled testing with known ground truth & Validity bounded by generator assumptions\\
\botrule
\end{tabularx}
\vspace{2pt}\par\scriptsize\textit{Abbreviations:} EHR, electronic health record; ICU, intensive care unit; ADNI, Alzheimer's Disease Neuroimaging Initiative; dVRK, da Vinci Research Kit.
\end{table}

\subsection{Evidence Layers and Translational Dimensions}\label{subsec:evidence_layers}

Across these application domains, the evidence base spans four
complementary layers. General world-model and embodied-intelligence studies
provide architectural principles, reasoning paradigms, and
environment-level evaluation methods
\cite{ref094,ref044,ref053,ref064,ref087,ref128}. Medical foundation
models, EHR trajectory models, and digital-twin research provide the
representational and clinical-integration background
\cite{ref085,ref111,ref003,ref062,ref110,ref117,ref119}.
Domain-specific studies contribute evidence from treatment simulation,
ultrasound guidance, surgical interaction, and population- or health-system
modelling
\cite{ref095,ref039,ref048,ref051,ref070,ref074,ref043,ref049,
ref052,ref073,ref083,ref084,ref125,ref127}. Reporting standards and
research on fairness, privacy, uncertainty, and safety define the
requirements for trustworthy evaluation and deployment
\cite{ref066,ref067,ref089,ref090,ref092,ref097,ref104,ref105,
ref107,ref108,ref109,ref114,ref123,ref124}.
Clinical translation, however, should be judged not by functional
capability alone, but by whether the available evidence supports the
intended clinical claim. We therefore organize the following translational
requirements around four interrelated dimensions: longitudinal data and
latent-state integrity, action semantics and causal validity,
uncertainty-aware evaluation and external validation, and safe deployment
and governance. Together, these dimensions provide an evidence-oriented
framework for assessing whether a medical world model should be interpreted
as a forecasting system, an intervention-conditioned simulator, a model for
comparing candidate interventions, or a candidate for prospective clinical
evaluation.

\subsection{Longitudinal Data and Latent-State Integrity}\label{subsec:data_integrity}

Large-scale EHR models demonstrate that longitudinal records can support patient-trajectory representation and longer-horizon prediction. Foresight, EHRWorld, ChronoMedicalWorld, MOTOR, EHRSHOT, CEHR-GPT, CoMET, TransformEHR, BEHRT, and Med-BERT cover complementary forms of timeline modelling and transfer \cite{ref085,ref047,ref054,ref007,ref015,ref019,ref023,ref041,ref100,ref106}. However, EHRs record care processes rather than the complete physiological state. A missing test can indicate clinical stability, limited access, loss to follow-up, or incomplete documentation; similarly, adherence, out-of-hospital medication, socioeconomic conditions, and disease severity are often only partially observed. A world model must therefore distinguish a stable state from an unmeasured change and from a measured but unstructured event. Multicentre EHR foundation models and TimelyGPT further show that portability across institutions, time horizons, and label-scarce settings cannot be assumed from scale alone \cite{ref111,ref059}.

Imaging-based treatment models face a different but related limitation: longitudinal samples are commonly small, disease-specific, and collected under restricted treatment protocols. MeWM used 338 paired single-centre CT scans before and after transarterial chemoembolization and 78 external cases, most of which contributed to adaptation and validation \cite{ref039}. TaDiff, CLARITY, and Brain-WM extend treatment-conditioned or long-term imaging simulation \cite{ref070,ref020,ref050}, but irregular follow-up, sparse treatment combinations, scanner variation, and patient heterogeneity constrain the states and transitions that can be learned. These datasets are appropriate for methodological exploration, but they do not establish reliable coverage of rare subtypes, delayed adverse events, or previously unseen treatment combinations.

Multimodal state construction adds temporal and semantic conflicts that simple feature concatenation cannot resolve. Imaging, laboratory tests, vital signs, pathology, genomics, and clinical text differ in sampling frequency, delay, resolution, and missingness mechanism. A unified latent state may conceal these conflicts unless the model represents observation time, data provenance, measurement uncertainty, and informative missingness explicitly. Future datasets should improve longitudinal coverage and intervention granularity, align modalities on clinically meaningful time scales, and report effective sample sizes at the patient, centre, and treatment-combination levels rather than only counts of images or events.

Several recent studies make these risks measurable. In a deliberately narrow evaluation of digital-biomarker forecasting, the error of a recurrent model rose by roughly 60\% when random-window splitting was replaced by patient-held-out evaluation, and exposure to another recording from the same patient improved apparent performance, illustrating how repeated-measure leakage can inflate generalization \cite{subramanian2026leakage}. Healthcare-process artefacts and information-availability errors can similarly inflate performance \cite{agniel2018biases,yang2026ehr2trace}. A further risk arises when generated trajectories are reused for training: indiscriminate training on model-generated data causes the tails of the original distribution to disappear \cite{shumailov2024collapse}, which is particularly concerning for rare subtypes and delayed adverse events. Generative models can also memorize and reproduce training examples, so synthetic patient trajectories are not automatically private \cite{carlini2023extracting}. Table~\ref{tab:threats} summarizes these and related threats to validity across the four translational dimensions, together with diagnostic checks.

\subsection{Action Semantics and Causal Validity}\label{subsec:action_causality}

Progress beyond state prediction depends on defining what constitutes an action. Xray2Xray treats projection angle as an acquisition action, EchoWorld uses probe-pose change, MeWM represents combinations of drugs and embolic materials, and Surgical Vision World Model infers latent actions from procedural video \cite{ref040,ref074,ref039,ref036}. These actions differ in clinical intent, granularity, observability, and risk. Latent actions can exploit unlabeled video, but a learned dimension may mix instrument motion, camera change, and tissue response. Explicit actions are easier to constrain and audit, yet their annotation is costly and their combinations grow rapidly.

Therapeutic actions require more than a drug or procedure label. Dose, route, timing, duration, sequence, combination therapy, stopping rules, organ function, contraindications, and drug interactions can all alter outcomes. Counterfactual recurrent networks and Causal Transformer accordingly represent treatment as a time-varying variable influenced by prior patient states \cite{ref006,ref012}, while individualized treatment-effect and causal machine-learning studies emphasize the need to specify interventions, outcomes, and follow-up windows \cite{ref013,ref075}. A practical direction is a hierarchical action ontology in which clinical intent, regimen and timing, and device-level control occupy separate layers, with dose ceilings, physical limits, guideline constraints, and human-handover conditions encoded as executable rules. The estimand and target trial should be defined before modelling so that vague historical behaviour is not converted into an unauditable action vector \cite{ref098}.

Causal validity remains the key distinction between forecasting and intervention support. Observational records reveal only the outcome under the treatment actually received, while treatment assignment reflects disease severity, patient preference, clinician experience, resource availability, and institutional workflow. A model can therefore reproduce historical conditional distributions while misrepresenting treatment effects. MeWM's agreement with recorded regimens may show that the model captures local practice, but does not establish that those regimens are optimal; similarly, Brain-WM's treatment-prediction accuracy may describe clinician behaviour rather than identify the action that improves outcome \cite{ref039,ref050}. Imaging models can also encode confounding signals unrelated to pathology, further weakening causal interpretation \cite{ref099}.

Generative realism cannot substitute for intervention realism. Visually convincing postoperative images or action-controllable surgical videos do not prove that drug-effect direction, tissue force, necrosis extent, or complication risk changes correctly when the action changes \cite{ref039,ref036,ref051}. 
Claims that a medical world model can support intervention comparison require stronger evidence, including target-trial emulation, explicit causal assumptions, sensitivity analyses, and measured-confounding adjustment, combined where possible with mechanistic constraints such as pharmacokinetics, tumour growth, imaging geometry, or tissue mechanics~\cite{ref098,ref099}. These methods cannot eliminate unmeasured confounding; multicentre prospective comparisons of model-assisted care with usual practice remain necessary before descriptive rollouts can support normative recommendations.

Recent methodological studies sharpen this point. Across eight datasets with real actions, the forecasting configurations with the lowest prediction error were at or below chance in mechanism consistency---that is, in whether shifting an action moved the forecast in a direction known a priori---on four of five datasets with declared mechanisms, whereas a directional-supervision loss improved consistency without changing error \cite{zhang2026mechanism}. Encoding the known sign of drug effects improved out-of-distribution accuracy in intensive care world models, while reversing those signs degraded it \cite{xu2026physioworld}. Formal audits further show that patient world models can agree on transition kernels and arm-specific risks yet disagree on the fraction of patients harmed by switching treatment, which is the counterfactual quantity most relevant to intervention-aware reasoning \cite{qi2026causalaudit}. Negative-control outcomes and exposures, E-value sensitivity analyses, and explicit positivity diagnostics provide practical tools for detecting such failures \cite{lipsitch2010negative,vanderweele2017evalue}.

\subsection{Uncertainty-Aware Evaluation and External Validation}\label{subsec:uncertainty_evaluation}

Long-horizon rollout compounds uncertainty because errors in state estimation become inputs to later transitions. Evaluation should therefore quantify both predictive uncertainty and uncertainty caused by missing data, model choice, action ambiguity, and distribution shift. Calibration must be assessed across horizons and clinically relevant subgroups, not only at the next time point. Models should detect out-of-distribution states, communicate uncertainty in forms clinicians can interpret, and abstain or request additional observations when confidence is insufficient \cite{ref097,ref109}. A single confidence score is inadequate if it does not identify which state variable, transition, or candidate action is unreliable.

The strength of clinical claims should increase only when supported by corresponding levels of validation.
Technical studies may report reconstruction, discrimination, trajectory similarity, or planning performance, but these metrics should be accompanied by temporal and geographic external validation, subgroup analysis, calibration, and failure-case reporting. TRIPOD+AI and MI-CLAIM provide foundations for transparent model reporting \cite{ref066,ref089}; DECIDE-AI addresses early prospective evaluation within clinical workflows \cite{ref114}; and SPIRIT-AI and CONSORT-AI specify how protocols and trial reports should describe AI-supported interventions \cite{ref105,ref104}. Applying these frameworks sequentially helps separate reproducible technical performance from workflow benefit and, ultimately, patient outcomes.

External validation should include institutions, scanners, documentation practices, and treatment policies that were not used for development or fine-tuning. Temporal validation is equally important because clinical practice, coding conventions, and patient populations change. Prospective studies should measure unsafe recommendations, abstention quality, clinician override, workflow burden, and downstream outcomes in addition to conventional accuracy. Capability levels such as L1--L4 must remain separate from evidence maturity: a model can perform closed-loop rollout in a benchmark while still lacking the validation required for clinical use.
Accordingly, increasingly advanced world-model capabilities should not be interpreted as increasingly strong clinical evidence unless the corresponding validation requirements are met.

Practical uncertainty tools are available but must be adapted to trajectories. Deep ensembles and Monte Carlo dropout approximate epistemic uncertainty \cite{lakshminarayanan2017ensembles,gal2016dropout}, and conformal prediction provides distribution-free coverage guarantees under exchangeability \cite{angelopoulos2023conformal}; conformal calibration of an action-conditioned world model has, for example, been used to set failure-detection thresholds for surgical robot policies \cite{huang2026fomofd}. Calibration should be assessed explicitly rather than inferred from discrimination \cite{vancalster2019calibration}, and regularizers have been proposed to reduce miscalibration and covariate-shift sensitivity in medical world models \cite{khan2026caltwin}. Uncertainty estimates can themselves fail: under 50\% temporal masking, one study found that forecast error rose sharply while Monte Carlo dropout variance fell \cite{subramanian2026leakage}.

The history of deployed clinical prediction models shows why external validation is indispensable. An external evaluation of a widely implemented proprietary sepsis model reported a hospitalization-level area under the receiver operating characteristic curve of 0.63, with the model missing 67\% of patients with sepsis despite generating alerts for 18\% of hospitalized patients \cite{wong2021epic}, and dataset shift after deployment can silently degrade performance \cite{finlayson2021datasetshift}. Risk-of-bias and reporting tools such as PROBAST+AI and CLAIM, together with TRIPOD+AI, should be applied to world-model components that make predictive claims \cite{moons2025probastai,mongan2020claim,ref066}, and the selection of offline policies should rely on off-policy estimators whose assumptions are reported \cite{thomas2016ope,tang2021modelselection}. Randomized evaluation of clinical AI nevertheless remains uncommon: a systematic review identified 41 randomized trials of machine-learning interventions, none of which adhered to all CONSORT-AI reporting items \cite{plana2022rct}.

\begin{figure}[!t]
\centering
\includegraphics[width=0.98\textwidth]{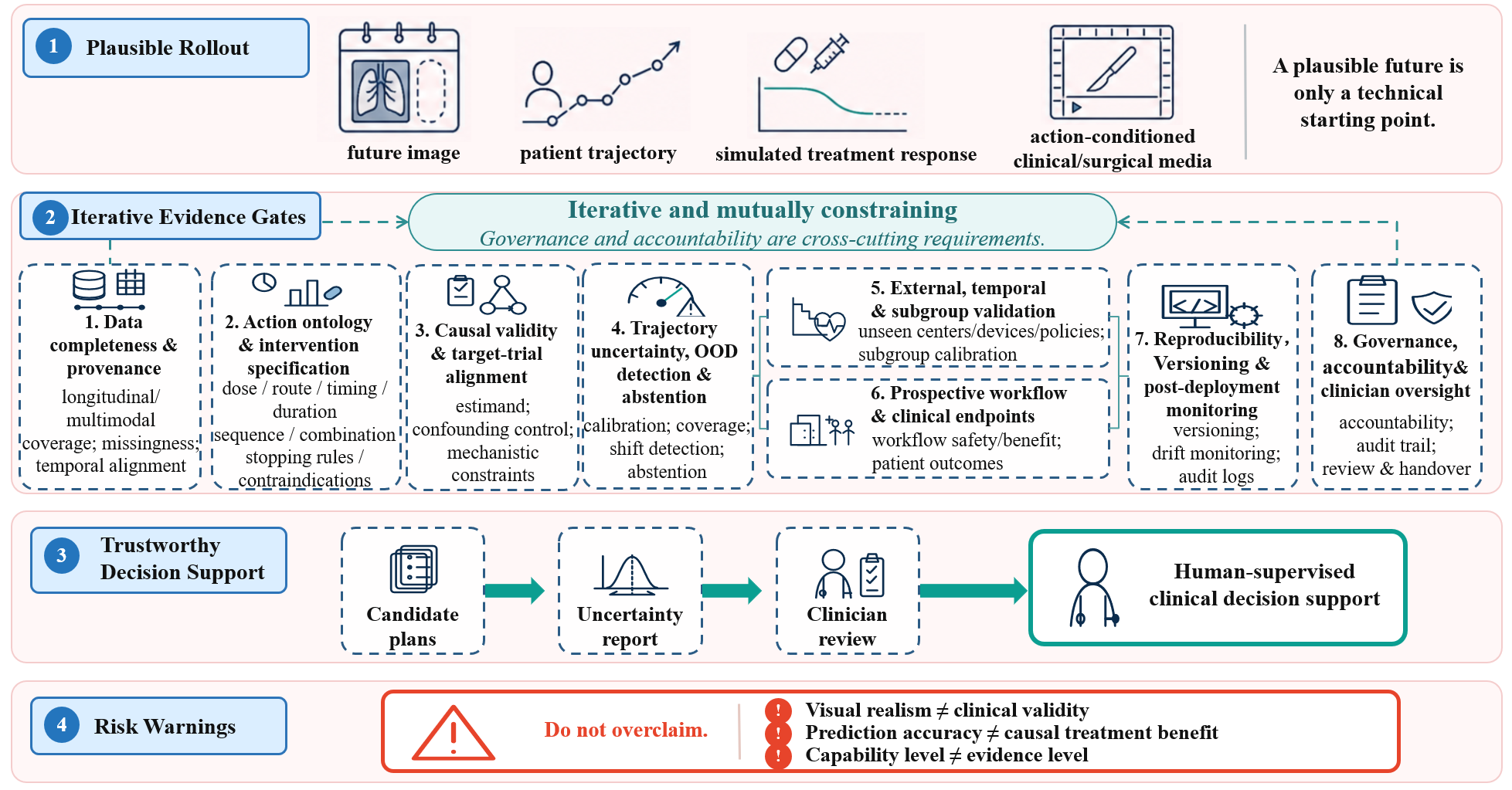}
\caption{Evidence framework for trustworthy clinical translation of medical world models. Model-generated outputs may include future imaging observations, patient trajectories, treatment-conditioned responses, and action-conditioned procedural media. Clinical interpretation requires evidence across data completeness and provenance, intervention specification, causal validity, trajectory uncertainty, external and prospective validation, clinically meaningful endpoints, reproducibility, post-deployment monitoring, governance, and clinician oversight. Candidate strategies should be accompanied by uncertainty estimates and clinician review. Visual realism, predictive performance, and functional capability alone do not establish clinical validity, causal treatment benefit, or evidence maturity.}\label{fig:evidence_chain}
\end{figure}

\subsection{Safe Deployment and Governance}\label{subsec:deployment_governance}

Deployment requires governance mechanisms that remain effective after initial validation. FUTURE-AI emphasizes trustworthy and deployable clinical AI, including fairness, traceability, robustness, and human oversight \cite{ref067}. These requirements are especially important for world models because errors can propagate across imagined trajectories and candidate plans. Performance should be audited across demographic and clinical subgroups, since medical AI can encode or infer protected attributes even when they are not explicitly supplied \cite{ref090,ref092}. Monitoring must also detect data drift, changes in action policies, and recurrent failure patterns after model updates.

Privacy protection must cover longitudinal linkage and multimodal patient states, which can be more identifying than isolated records. Federated learning can support multi-institutional development without centralizing raw data, but it does not by itself resolve heterogeneous labels, site-specific bias, leakage through model updates, or responsibility for downstream errors \cite{ref107,ref108}. Deployment agreements should define data provenance, access control, versioning, audit logs, update approval, incident reporting, and the institutions responsible for corrective action.

Finally, safety must be implemented as system behaviour rather than described only in prompts or policy statements. Safety-oriented analyses of world models highlight risks from model misspecification, reward or objective errors, distribution shift, and overconfident planning \cite{ref123,ref124}.
Before planning-oriented outputs are interpreted as clinical decision support, medical world models should provide auditable candidate trajectories and plans for clinician review rather than autonomous prescriptions. They require explicit stop conditions, uncertainty-triggered abstention, guideline and physical constraints, and reliable handover to human decision-makers. 
Figure~\ref{fig:evidence_chain} summarizes this progression from technically plausible rollout to a clinically governed evidence chain.

Lessons from deployed clinical AI apply directly. The number of AI-enabled medical devices cleared by the US Food and Drug Administration has grown rapidly \cite{benjamens2020fda}, yet most were evaluated only retrospectively at the time of submission \cite{wu2021fdadevices}. Successful integration into routine care requires sustained investment in workflow design, stakeholder engagement, and staff training \cite{sendak2020sepsiswatch,kelly2019keychallenges}, and clinicians can over-rely on automated advice, a phenomenon known as automation bias \cite{goddard2012automation}. Algorithms that optimize proxies can encode inequity, as shown by a widely used population-health algorithm that underestimated the needs of Black patients because it predicted health-care costs rather than illness \cite{obermeyer2019bias}; world models that optimize simulated outcomes face the same risk when objectives are mis-specified. Federated learning has enabled multi-institutional clinical models without centralizing raw data \cite{mcmahan2017fedavg,dayan2021exam}, but the first benchmark of federated clinical world models found that client ownership and participation restricted the coverage of long trajectory windows, and that algorithmic gains were small and horizon-dependent \cite{wei2026federatedwm}. Runtime monitors based on world-model predictions, such as conformal failure detectors, and formal safety shields offer further safeguards during deployment \cite{huang2026fomofd,alshiekh2018shield}.

\begin{table}[!t]
\scriptsize
\setlength{\tabcolsep}{2pt}
\renewcommand{\arraystretch}{1.05}
\caption{Common threats to the validity of medical world models, how they arise, and diagnostic checks or mitigations. Threats are grouped approximately by the four translational dimensions: longitudinal data and latent-state integrity; action semantics and causal validity; uncertainty-aware evaluation and external validation; and safe deployment and governance.}\label{tab:threats}
\begin{tabularx}{\textwidth}{@{}>{\raggedright\arraybackslash}p{0.17\textwidth}YY>{\raggedright\arraybackslash}p{0.16\textwidth}@{}}
\toprule
Threat & How it arises in medical world models & Diagnostic check or mitigation & Sources\\
\midrule
Temporal and information-availability leakage & Events are time-stamped by reference rather than availability, so future information enters the history & Availability-filtered histories; prospective-style temporal splits & \cite{yang2026ehr2trace,agniel2018biases}\\
Patient-level and repeated-measure leakage & Windows from the same patient or recording appear in training and test sets & Patient-held-out and recording-excluded splits & \cite{subramanian2026leakage}\\
Informative observation & Measurement depends on the suspected state & Explicit observation models; sensitivity to missingness mechanisms & \cite{agniel2018biases,che2018grud}\\
Confounding by indication & Sicker patients receive more treatment, reversing apparent effects & Target-trial emulation; known-sign constraints; negative controls & \cite{hernan2016target,xu2026physioworld,lipsitch2010negative}\\
Positivity violation & Queried strategies are rarely or never observed for similar patients & Overlap diagnostics; support-restricted queries & \cite{robins2000msm,gao2026camewm}\\
Action misspecification & Single components are edited outside realistic bundles, or latent actions mix several factors & Bundle-aware editing; hierarchical action ontologies & \cite{wang2026bundles,ref036}\\
Action unfaithfulness & Accurate forecasts are insensitive to, or wrongly signed for, action changes & Mechanism-consistency and action-following tests & \cite{zhang2026mechanism,yang2026sonoworld}\\
Compounding rollout error & Errors in early transitions propagate to later states & Horizon-resolved metrics; drift-aware training; frequent replanning & \cite{pan2026surgvista,yang2026clinjepa,li2026sonograph}\\
Miscalibration under shift & Long-horizon forecasts are overconfident at new sites or times & Horizon- and subgroup-specific calibration; external validation & \cite{vancalster2019calibration,khan2026caltwin,wong2021epic}\\
Generative hallucination & Images or videos are realistic but clinically incorrect & Lesion-level reader studies; clinical checklists & \cite{cohen2018hallucinate,tan2026remedy,ref083}\\
Shortcut learning & Site, device, or acquisition markers drive predictions & Stress tests across sites and devices & \cite{geirhos2020shortcut,degrave2021shortcuts}\\
Spurious personalization & A population-average forecast is presented as patient-specific & Patient, place, and prior ablation audits & \cite{gu2026p3}\\
Recursive synthetic training & Training on generated trajectories erodes distribution tails & Provenance tracking; validation on real data & \cite{shumailov2024collapse}\\
Inequitable objectives & Proxy rewards misrepresent need & Subgroup audits of objectives and outcomes & \cite{obermeyer2019bias,ref090}\\
Over-reliance & Clinicians defer to confident rollouts & Interpretable uncertainty displays; human-factors evaluation & \cite{goddard2012automation,ref114}\\
\botrule
\end{tabularx}
\end{table}


\section{Open Problems and Future Directions}\label{sec:future}

The preceding analysis suggests that progress will depend less on larger generators than on tighter coupling between models, causal design, mechanistic knowledge, and evaluation. We highlight eight directions, summarized in Table~\ref{tab:agenda}.

\subsection{Causal--mechanistic hybrid world models}

Purely data-driven transitions learn the conditional distributions present in historical care, whereas mechanistic models encode relationships that are expected to remain valid under intervention. Hybrid designs that integrate pharmacokinetic and pharmacodynamic models, tumour-growth equations, cardiac electrophysiology, or known effect directions into latent transitions are promising because they constrain extrapolation to unobserved actions \cite{qian2021expertode,chen2026ecgwm,xu2026physioworld,gillette2021cardiactwin}. Their evaluation should test not only predictive accuracy but also whether the encoded mechanisms are respected under changed actions \cite{zhang2026mechanism}, and whether the causal estimand targeted by the rollout is identified \cite{gao2026camewm,hernan2016target}.

\subsection{Multiscale and continuous-time patient states}

Clinical processes unfold on time scales ranging from milliseconds in cardiac electrophysiology to decades in chronic disease, and on spatial and organizational scales ranging from molecules to populations. Continuous-time latent models and event-driven updates can accommodate heterogeneous sampling \cite{rubanova2019latentode,kidger2020ncde,chung2026afablation}, and multiscale models, such as an immune world model spanning cellular to individual levels, suggest how hierarchical states might be composed \cite{cui2026immunewm,laubenbacher2022immune}. A key open question is how to connect fast procedural states with slow disease states, so that, for example, intraoperative actions can be linked to long-term outcomes.

\subsection{Hierarchical clinical action ontologies}

Medical actions span clinical intent, regimen and timing, and device-level control. A shared hierarchical action vocabulary, analogous to the common data models used for observations, would allow treatments to be represented as coherent bundles with doses, routes, and durations, and would make action conditioning auditable across institutions \cite{wang2026bundles,hripcsak2015ohdsi}. Language-conditioned actions offer flexibility \cite{ref051,he2025cosmoshsurgical}, but they must be grounded in executable and measurable quantities.

\subsection{Uncertainty-aware and safe planning}

Planning with learned models should be conservative by design, penalizing rollouts outside the support of the data, enforcing hard constraints, and abstaining when uncertainty is high \cite{yu2020mopo,kidambi2020morel,achiam2017cpo,alshiekh2018shield}. Medical applications additionally require that uncertainty be communicated in clinically interpretable terms and that abstention trigger a defined handover to clinicians \cite{ref109}. Runtime monitors that compare observed and predicted transitions may provide an additional layer of protection during execution \cite{huang2026fomofd}.

\subsection{World models as substrates for clinical agents}

Language-model agents are increasingly evaluated in simulated clinical environments \cite{schmidgall2024agentclinic,jiang2025medagentbench,li2024agenthospital}. Coupling such agents with learned patient world models could allow them to test candidate actions before making a recommendation \cite{wu2026sepsisagent}, and language models themselves may encode partial knowledge of clinical dynamics \cite{willems2026futurequerying}. This coupling, however, compounds errors across the agent and the world model, and it requires joint evaluation of reasoning, simulation, and action.

\subsection{Benchmarks, simulators, and evaluation science}

The field lacks benchmarks that test counterfactual and planning claims rather than reconstruction or forecasting alone. Promising elements include action-following and mechanism-consistency tests, causal harm bounds, personalization audits, semi-synthetic environments with known ground truth, and motion-centric surgical benchmarks \cite{zhang2026mechanism,qi2026causalaudit,gu2026p3,liu2026surgwmbench,ref064}. Simulators used for evaluation must themselves be validated for the specific question they are asked to answer, as exemplified by the purpose-restricted regulatory acceptance of the UVA/Padova simulator \cite{kovatchev2009insilico}.

\subsection{Privacy-preserving and federated world models}

Longitudinal, multimodal patient states are highly identifying, and individual trajectories are often partitioned across institutions. Federated training must therefore address not only data heterogeneity but also fragmented temporal coverage of trajectories \cite{wei2026federatedwm,ref108}, and generated trajectories should be assessed for memorization before release \cite{carlini2023extracting}.

\subsection{Prospective evaluation and lifecycle governance}

Claims of clinical benefit ultimately require prospective evaluation. Staged studies that follow DECIDE-AI, SPIRIT-AI, and CONSORT-AI, with predefined stopping rules and monitoring for drift, are needed before planning-oriented outputs influence care \cite{ref114,ref105,ref104}. Because world models will be updated as data accumulate, governance should define change-control procedures, revalidation triggers, and accountability for model-informed recommendations \cite{ref067,nasem2024digitaltwins}.

\begin{table}[!t]
\scriptsize
\setlength{\tabcolsep}{2pt}
\renewcommand{\arraystretch}{1.08}
\caption{Research agenda for medical world models. For each direction, the table lists the principal current gap, near-term research priorities, and an evidence milestone that would indicate meaningful progress towards clinical use.}\label{tab:agenda}
\begin{tabularx}{\textwidth}{@{}>{\raggedright\arraybackslash}p{0.17\textwidth}YYY@{}}
\toprule
Direction & Current gap & Near-term research & Evidence milestone\\
\midrule
Causal--mechanistic hybrids & Transitions reproduce historical care patterns & Embed pharmacological, physiological, and growth constraints; target prespecified estimands & Correct effect directions under changed actions and agreement with randomized evidence\\
Multiscale, continuous-time states & Fixed time steps and single-scale states & Event-driven and continuous-time latent models; hierarchical states & Calibrated forecasts across horizons from hours to years\\
Hierarchical action ontologies & Actions are vague, latent, or isolated components & Shared vocabularies for intent, regimen, and device control; bundle-aware editing & Auditable action conditioning that transfers across institutions\\
Uncertainty-aware safe planning & Planners exploit model errors & Support-constrained planning; hard constraints; abstention and handover & Zero critical constraint violations in staged evaluation\\
Agents coupled with world models & Agents lack grounded patient dynamics & Propose--simulate--refine workflows with joint evaluation & Prospective evaluation of agent recommendations with clinician oversight\\
Benchmarks and evaluation science & Metrics reward realism and forecasting & Counterfactual, mechanism-consistency, and personalization benchmarks & Community benchmarks with known ground truth for L3 and L4 claims\\
Privacy and federation & Trajectories fragmented across institutions; memorization risk & Trajectory-aware federated learning; privacy auditing of generated data & Multi-institutional models with documented privacy guarantees\\
Prospective lifecycle governance & Few prospective or randomized studies & Staged evaluation; change control; post-deployment monitoring & Prospective multicentre studies with clinically meaningful endpoints\\
\botrule
\end{tabularx}
\end{table}

\subsection{Limitations of this Review}\label{subsec:limitations}

This Review has several limitations. The database searches were completed in July 2026, and screening was machine-assisted and author-verified rather than based on exhaustive dual screening. The field is evolving rapidly; the directed identification of recent studies during manuscript development partly addresses this, but many recent studies are preprints whose findings may change during peer review. The L1--L4 framework describes functional capability and may not capture all relevant distinctions, such as those between therapeutic and non-therapeutic actions or the status of world--action models that combine prediction and control. Finally, because the included studies are heterogeneous in tasks, data, and outcomes, we did not attempt quantitative synthesis, and our appraisal of evidence maturity is qualitative.

\section{Conclusion}\label{sec:conclusion}

Medical world models extend medical AI beyond static prediction towards longitudinal state modelling, intervention-conditioned simulation, and planning-oriented decision support. Current evidence demonstrates technical feasibility across medical imaging, longitudinal patient modelling, treatment-response simulation, embodied clinical interaction, and population-level health modelling, but remains insufficient to establish clinical effectiveness. Our synthesis indicates that stronger clinical claims require correspondingly stronger evidence: temporal forecasting requires robust longitudinal and external validation; intervention-conditioned simulation requires explicit action semantics, uncertainty quantification, and causal justification; alternative-intervention comparison requires explicit causal estimands, defensible causal assumptions, and validation against clinically meaningful outcomes; and planning-oriented decision support further requires prospective evaluation, safety constraints, and clinician oversight. Functional capability should therefore remain distinct from clinical evidence maturity. The key translational challenge is not simply to generate plausible future trajectories, but to determine when they are sufficiently reliable, interpretable, and clinically validated to inform decision making while preserving clinician responsibility.

Looking ahead, the most consequential advances are likely to come from hybrid models that respect clinical mechanisms and causal structure, from action representations that mirror how care is delivered, and from evaluation frameworks that test whether models respond correctly to interventions rather than merely reproducing historical data. Achieving these goals will require close collaboration among machine-learning researchers, clinicians, epidemiologists, and regulators, so that medical world models are developed and evaluated as tools for generating clinical evidence rather than as stand-alone generative systems.
\newpage
\section*{Declaration statements}

\subsection*{Funding}
This work was supported in part by the National Natural Science Foundation of China (No. 62502064), Joint Plan of Liaoning Province Science and Technology Plan (No. 2025JH2/101800417), Scientific Research Project of Liaoning Provincial Department of Education (No. LJ222511258003), Joint Plan of Liaoning Province Science and Technology Plan (No. 2025JH2/101800422), Interdisciplinary Project of Dalian University (No. DLUXK-2025-QN-020), 111 Center (No. D23006).

\subsection*{Conflict of interest/Competing interests}
The authors declare no competing interests.

\subsection*{Ethics approval and consent to participate}
Not applicable. This review does not involve new experiments on humans, animals, or human samples.

\subsection*{Consent for publication}
Not applicable.

\subsection*{Data availability}
No new datasets were generated or analysed in this review.

\subsection*{Materials availability}
Not applicable.

\subsection*{Code availability}
Not applicable.

\newpage
\subsection*{Author Contributions}

Wanshu Fan led the conceptualization of the study, defined the topic, scope,
and overall framework of the review, coordinated the evidence synthesis, and
was primarily responsible for manuscript drafting, integration, and revision.
Zhongxiu Cong and Zhaoyan Chen contributed to the conceptual development of
the review and jointly designed and conducted the literature search and
evidence-synthesis process, including database searching, duplicate removal,
title and abstract screening, full-text assessment, reference verification,
and evidence-table preparation. They also contributed to manuscript writing,
literature verification, and the preparation and revision of figures and
tables. Zhuanfeng Jin contributed to literature screening and organization,
evidence verification, preparation of supporting materials, and manuscript
revision.

Qi Ai, Haifan Gong, Congyu Liao, and Xiaofeng Liu provided expert input from
the perspectives of clinical medicine, medical artificial intelligence, and
related technical fields. They reviewed the medical concepts, technical
classification, representative studies, clinical applications, and
translational and trustworthy-AI considerations discussed in the manuscript,
and provided critical comments on the academic content.

Dongsheng Zhou and Cong Wang jointly supervised the study and provided
guidance on the review methodology, evidence organization, manuscript
structure, and interpretation of the findings. They critically revised the
manuscript for important intellectual content and reviewed the consistency
between the cited evidence and the manuscript conclusions. All authors
reviewed and approved the final version of the manuscript and agreed to be
accountable for the content of the work.
\newpage
\bibliography{medical_world_model_refs}

\end{document}